\documentclass[preprint,11.5pt]{elsarticle}

\usepackage{amssymb}
\usepackage{algorithmic}
\usepackage{algorithm}
\usepackage{multirow}
\usepackage{makecell}
\usepackage{array}
\usepackage{amsmath,amsfonts}
\usepackage{booktabs}
\usepackage{color}
\usepackage{enumitem}
\usepackage{hyperref}
\usepackage{setspace}

\usepackage{geometry}
\journal{Decision Support Systems}

\begin{document}

\begin{frontmatter}
\begin{spacing}{2.0}


\title{Efficient Fraud Detection Using Deep Boosting Decision Trees}

 \author[inst1]{Biao Xu}
 \ead{xubiao.xjtu@gmail.com}

 \address[inst1]{{School of Management, Center of Intelligent Decision Making and Machine Learning},
             {Xi'an Jiaotong University}, 
             {Xi'an},
             {710049}, 
             {Shanxi},
             {P.R.China}}
 \author[inst1]{Yao Wang}
 \ead{yao.s.wang@gmail.com}

 \author[inst1]{Xiuwu Liao}
 \ead{liaoxiuwu@mail.xjtu.edu}
 \author[inst1]{Kaidong Wang\corref{correspondingauthor}}
 \cortext[correspondingauthor]{Corresponding author at: School of Management, Xi'an Jiaotong University, No.28, West Xianning Road, Xi'an, Shaanxi, 710049,P.R. China.}
 \ead{wangkd13@gmail.com}

\begin{abstract}

Fraud detection is to identify, monitor, and prevent potentially fraudulent activities from complex data. The recent development and success in AI, especially machine learning, provides a new data-driven way to deal with fraud. From a methodological point of view, machine learning based fraud detection can be divided into two categories, i.e., conventional methods (decision tree, tree boosting methods...) and deep learning, both of which have significant limitations in terms of the lack of representation learning ability for the former and interpretability for the latter. Furthermore, due to the rarity of detected fraud cases, the associated data is usually imbalanced, which seriously degrades the performance of classification algorithms.

In this paper, we propose deep boosting decision trees (DBDT), a novel approach for fraud detection based on gradient boosting and neural networks. In order to combine the advantages of both conventional methods and deep learning, we first construct soft decision tree (SDT), a decision tree structured model with neural networks as its nodes, and then ensemble SDTs using the idea of gradient boosting. In this way we embed neural networks into gradient boosting to improve its representation learning capability and meanwhile maintain the interpretability. Furthermore, aiming at the rarity of detected fraud cases, in the model training phase we propose a compositional AUC maximization approach to deal with data imbalances at algorithm level. Extensive experiments on several real-life fraud detection datasets show that DBDT can significantly improve the performance and meanwhile maintain good interpretability. Our code is available at \href{https://github.com/freshmanXB/DBDT}{https://github.com/freshmanXB/DBDT}.
\vspace{1em}
\end{abstract}



\begin{keyword}
fraud detection \sep boosting \sep deep learning \sep compositional AUC maximization

\end{keyword}
\end{spacing}
\end{frontmatter}


\section{Introduction}
\label{sec:Intro}
\begin{spacing}{2.0}
\vspace{-2mm}
In modern society fraud exists in all walks of life and causes serious economical losses to society and individuals \cite{bao2022artificial}. For example, IBM released 2022 IBM Global Financial Fraud Impact Report, which revealed that global consumers have moved nearly exclusively to credit card and digital payments \cite{RN12}, and losses associated with credit card fraud, a common type for fraud, in the United States alone are close to \$17 billion \cite{AnnaMaria}. This is just the tip of the iceberg. Actually losses due to fraudulent activities keep increasing each year, causing worldwide adverse effects \cite{kou2021fintech}. Therefore, fraud detection, as a means to detect and prevent fraudulent activities in time, has been more important than ever before and aroused widespread concern. In the early days, researches on fraud detection has mainly focused on rule-based expert system embedding specific domain knowledge \cite{leonard1995development}. More recently, the development and success in artificial intelligence, especially machine learning, has opened potential new venues to tackle fraud \cite{bao2022artificial}.

From a machine learning perspective, the existing researches usually treat fraud detection as a binary classification problem and train prediction models using off-the-shelf supervised machine learning algorithms, such as CART \cite{breiman2017classification} , XGBoost \cite{chen2016xgboost} and SMOTBoost \cite{chawla2003smoteboost}. In addition to those, deep learning based on neural networks has achieved great success in many applications owing to its powerful representation learning capability, and thus has also been introduced into this field \cite{zhang2021hoba, sanober2021enhanced}. Although having been shown to hold promise, machine learning, especially deep learning, still encounters some challenges in fraud detection task. 

The first is about the choice between conventional methods and deep learning, both of which have their advantages and disadvantages. Many conventional machine learning methods have good interpretability (such as the clear decision path for decision tree and the feature importance for boosting model), but has shortcomings in data representation and generalization. And on the contrary, deep learning techniques are capable of effectively learning data representations and exhibiting superior generalization performance, while it is known that deep learning is a black box, leaving managers perplexed about their underlying mechanisms.
\textit{Therefore, how to strike a balance between the two methodologies is an issue to consider}. 

The second, due to the rarity of detected fraud cases, the associated training data is usually vary imbalanced, which seriously degrades the performance of classification algorithms. This is because most existing algorithms are designed to maximize classification accuracy and reduce error, and thus work best when the number of samples in each class is approximately equal. \textit{Thus, how to deal with the data imbalance in fraud detection is the key to improve the algorithm performances.} The aforementioned two issues in fraud detection are exactly what we aim to address in this paper.

Concretely, our work contributes to the literature by proposing a novel fraud detection algorithm named deep boosting decision trees (DBDT) which gives full play to the advantages of both conventional machine learning methods and deep learning techniques, and meanwhile effectively handles imbalanced data at algorithm level (rather than under- or over-sampling the data \cite{piri2018synthetic,tanha2020boosting}). Our basic starting point is to embed neural networks into gradient boosting decision trees, a representative conventional learning method, to improve its representation learning capability and meanwhile maintain the basic forward structure (and thus its interpretability can be fully preserved). Furthermore, compared with accuracy and misclassification error, AUC has proven to be a more reasonable performance evaluation metric for fraud prediction models \cite{cortes2003auc}, and thus we consider to employ a AUC maximization approach for the DBDT optimization to deal with data imbalance at algorithm level. 

Essentially, there are three main challenges we met in the DBDT construction. \textit{First, how to construct differentiable decision tree embedding neural networks.} As in gradient boosting decision trees, we also employ decision tree-like structure as the base learner in DBDT. In the conventional decision tree model like ID3 \cite{quinlan1986induction}, C4.5 \cite{quinlan2014c4}, CART \cite{breiman2017classification}, the split criterion is greedy, which is particularly affected by noise and might be erratic. Therefore, decision trees do not usually generalize as well as deep neural network \cite{frosst2017distilling}, and it is easy to overfitting. Thus, it is the first thing for us to learn split criterion in deep learning framework to enhance the generalization of decision tree and make it more soft and stable. \textit{Second, how to ensemble the differentiable decision trees in the gradient boosting framework.} We know that gradient boosting decision trees are serial structure meaning that the construction of the current decision tree depends on the result of the previous one, which is a natural contradiction with the end-to-end parallel training framework of deep learning. Therefore, based on the differentiable decision trees, what next is to concatenate them together to get a gradient boosting decision trees like end-to-end forward structure (DBDT) which can be jointly optimized. \textit{Finally, how to design stochastic algorithms with provable guarantees to optimize the resulting model in the sense of AUC maximization.} We know that in general a neural network model for classification tasks can be trained by minimize the misclassification rate and its corresponding surrogate losses \cite{rosenblatt1961principles,rumelhart1985learning}, in which cross-entropy (CE) loss a typical example. While in the AUC case, things will be different. Actually, compared with the canonical neural networks, conducting end-to-end training for DBDT is an inherently different and more complex problem, and optimizing an AUC surrogate loss from scratch directly does not yield a cleaner feature representations for fraud and non-fraud classes of data than optimizing the cross-entropy loss \cite{yuan2021compositional}. Thus, maximizing an AUC surrogate loss directly does not yield a satisfactory performance. Taking that into consideration, when training DBDT for fraud detection it is an important issue to maximize AUC meanwhile avoid feature representation degradation.

The main thread of this work is to address the above problems. Our model refrains from directly modeling decision tree using greedy algorithm for the reasons mentioned earlier, but instead employ soft decision tree (SDT) as the base learner. We build SDT as a binary tree structure using neural networks as the routing gate within the inner nodes to get a better data representation and generalization, and the final prediction for the input sample is determined by the weighted sum of class distributions among all leaf nodes. Then we accomplish DBDT by integrating SDTs to a end-to-end structure using a global loss to force each SDT to fit the residuals from the previous step, where the residuals are calculated by the negative gradients. Apparently DBDT is ideologically in common with gradient boosting decision trees, while in DBDT those SDTs can be jointly parallel optimized. Finally, we cast imbalanced DBDT optimization into a non-convex concave min-max stochastic AUC optimization problem \cite{liu2019stochastic} and apply compositional deep AUC maximization algorithm to solve our model for fraud detection. As we will see later, in the training phase we minimize a two-level compositional objective, where the inner ensures our model have basic tree boosting construction and classification ability, and the outer further guarantees a excellent performance in imbalanced fraud data. In this way, DBDT realizes the effective processing of imbalanced data at the algorithm level as opposed to directly modeling after re-sampling.

In an extensive experimental evaluation, DBDT is shown to outperform relevant benchmarks on various real-life fraud data. Furthermore, we study the parameter sensitivity, including the depth of SDT, the depth of neural networks in SDT and the number of SDTs for ensembling, and the results show that DBDT is relatively stable for hyperparameters and is not easy to overfit.  More specifically, the case studies demonstrate that DBDT maintains
good interpretability.

In summary, the main contributions of this research are:

(1) We propose a novel fraud detection model named deep boosting decision trees (DBDT). In order to give full play to the advantages of both conventional machine learning methods and deep learning, DBDT embeds neural network into gradient boosting machine, which significantly improves its representation learning capability and meanwhile maintains the interpretability. 

(2) Aiming at the rarity of detected fraud cases, in the model training phase we propose a compositional AUC maximization approach to deal with data imbalances at algorithm level. Experiments show that this optimization strategy can significantly improve the performances in fraud detection.

(3) Extensive experiments on various types of real-life fraud data demonstrate that our model outperforms relevant benchmarks and meanwhile maintains good interpretability.

The rest of the paper is organized as follows. Section 2 is a review of related work. In Section 3, we describe our model specifications. Section 4 and 5 describe the benchmarks and the experimental design respectively, before the results are presented in Section 6. We conclude the paper in Section 7.
\end{spacing}

\vspace{-4mm}
\section{Related Work}
\label{sec:Related}
\begin{spacing}{2.0}
\vspace{-2mm}
\subsection{Fraud Detection}
Fraud detection originally relied on rule-based expert system embedding specific domain knowledge \cite{leonard1995development}, and more recently machine learning has been applied most extensively to mine fraudulent patterns \cite{ngai2011application}. Earlier research on fraud detection focused on classification techniques such as Logistic Regression, SVM and Bayesian Belief Networks, as well as neural networks \cite{hoogs2007genetic,yue2007review,quah2008real}. While they are widely used, these classical methods do not account for why they can make sense in imbalanced data and need to collect fraud activities frequently and relearn periodically. Actually, several studies have shown that Logistic Regression, SVM and Random Forests all perform significantly better at detecting legitimate transactions correctly than fraudulent ones \cite{dhankhad2018supervised}. However, fraud detection is a problem with a large difference in misclassification costs: it is typically far more expensive to misdiagnose a fraudulent transaction as legitimate than the reverse \cite{west2016intelligent}. 

tree boosting method can attach more attention to fraud cases, and several recent studies have shown that the use of tree boosting method can detect fraud well. A representative work in this area is by Viaene et al. \cite{viaene2004case}, which applies the weight of evidence reformulation to AdaBoosted naive Bayes scoring in the problem of diagnosing insurance claim fraud. Since then, there has been much follow-up works on applying boosting algorithm to fraud detection. More recently, Yang Bao et al. \cite{bao2020detecting} have used RUSBoost to demonstrate the value of combining accounting fraud domain knowledge and machine learning methods in model building. There has also been works on applying neural networks to fraud detection, e.g., node representation learning \cite{van2022catchm}, graph neural networks \cite{xu2021towards, van2015apate} and recurrent neural network \cite{lin2021online}. Though machine learning based fraud detection is methodologically elegant and can effectively predict fraudulent activities, these approaches are weaker generalization which based on tree boosting algorithms and worse interpretable which based on neural networks. Specifically, some tree boosting algorithms deal with the data imbalances depending on re-sampling unlike our approach which can get better performance applying original data.

\vspace{-2mm}
\subsection{Deep Learning for Boosting}
Fraud is adversarial \cite{craja2020deep}. In the case of fraud, perpetrators try to prevent the learning \cite{bao2022artificial}, and thus supervisors need more generalization when applying machine learning method to fraud detection. Past researches have focused on accuracy and interpretability in tree boosting method for fraud detection, but ignore generalization, which becomes more and more important in the era of credit card and digital payments. Deep learning based on neural networks has achieved great success in many fields owing to its powerful representation learning capability, and thus can be employed to improve the generalization of boosting. The most straightforward way is to retrain a decision tree (DT) using deep learning framework and then ensemble all DTs by boosting algorithms. One of the earliest works in this area was by Frosst et al. \cite{frosst2017distilling} which distill a neural network into a soft decision tree (SDT). This research takes the knowledge acquired by the neural net and expresses the same knowledge in a model that relies on hierarchical tree-like decisions instead, in which explaining a particular decision would be much easier.  More recently, Chih-Hsuan Wang et al. \cite{wang2016integrating} integrate decision tree with back propagation network to conduct business diagnosis and performance simulation for solar companies. Alvin Wan et al. \cite{wan2020nbdt} propose NBDTs replacing a neural network's final linear layer with a differentiable sequence of decisions and a surrogate loss, which extends SDT to jointly improve both the accuracy and interpretability.

\vspace{-4mm}
\subsection{Compositional Training for Deep AUC Maximization}
When applying neural networks to balanced data such as image classification task \cite{simonyan2014very,lecun1998gradient}, we usually employ a surrogate loss of the prediction error (e.g., CE loss), and compare the predictions with the corresponding ground-truth labels \cite{ioffe2015batch,szegedy2015going}. While in the case of badly imbalanced data such as fraud detection task, AUC is a commonly used alternative to prediction error for the  evaluation of model performance, and thus we need a surrogate loss of the AUC. More recently, AUC maximization problem is widely concerned \cite{ying2016stochastic,natole2018stochastic,liu2019stochastic,yuan2021compositional}. For example, the most earliest work in AUC maximization suggests that the performance of a learner by classification accuracy is inappropriate for applications where the data are unevenly distributed among different classes, and presents two algorithms for online AUC maximization with theoretic performance guarantee. Another line of research has focused on formulating AUC optimization as a convex-concave saddle point problem and solve it using stochastic online algorithm. Recent research has also focused on the degraded feature representations learned by maximizing the AUC loss from scratch \cite{yuan2021compositional}. Compositional training for deep AUC maximization is a better and novel end-to-end training method that achieves both benefits of minimizing the exponential loss for robust feature learning and minimizing an AUC loss for robust classifier learning \cite{yuan2021compositional}. 
\end{spacing}

\vspace{-4mm}
\section{Deep boosting decision trees for fraud detection}
\begin{spacing}{2.0}
In this section we introduce the construction and optimization of our DBDT model for fraud detection in greater detail, which mainly involves three stages: the construction of soft decision tree, integrating soft decision trees using gradient boosting, and the compositional AUC maximization approach for fraud detection. 
\vspace{-2mm}
\subsection{Construction of Soft Decision Tree}
We build soft decision tree (SDT) as a binary tree structure, and to improve its representation and generalization, we build each node in SDT as a multilayer perceptron with several hidden layers. All nodes of a decision tree can be divided into two types: inner nodes and leaf nodes.
\vspace{-2mm}
\subsubsection{Generating Inner Nodes}
\indent
In SDT, each inner node is designed to be a multilayer perceptron whose input is the sample
features and output the probability of selecting the right sub-path, and a sigmoid activation function $\sigma(\boldsymbol{x})=(1+\exp (-\boldsymbol{x}))^{-1}$ is used to bring to nonlinearity and enhance the representation of the SDT. Thus, the input dimension of the inner node network is the feature dimension and the output dimension is 1.
Taking the threshold as $0.5$, we then compare the probability of selecting the right sub-path $p_{i}$ with the threshold. When the probability of the right sub-path $p_{i} \geqslant 0.5$, the right path is selected, and when $p_{i} < 0.5$, the left sub-path is selected. In our paper, we indicate $\mathbb{I}_{i}^{\ell}$ and $\mathbb{I}_{i}^{r}$ respectively as  whether the node selects the left path or right path. $\mathbb{I}_{i}^{\ell}$ and $\mathbb{I}_{i}^{r}$ satisfy $\mathbb{I}_{i}^{\ell}+\mathbb{I}_{i}^{r}=1$, and can be defined as
\begin{spacing}{1.3}
\begin{equation}
\setlength{\abovedisplayskip}{5.5pt}
\setlength{\belowdisplayskip}{5.5pt}
\mathbb{I}_{i}^{\ell}= \begin{cases}1 & p_{i}<0.5 \\ 0 & p_{i} \geqslant 0.5.\end{cases}
\end{equation}
\end{spacing}

\vspace{-2mm}
\subsubsection{Generating Leaf Nodes}
\indent
Unlike the inner node described above, the output dimension of leaf node depends on our machine learning task. For instance, when we using SDT to deal with classification task, the output dimension of the leaf node is classes $K$. For fraud detection, it means $K=2$. At the same time, the output of each leaf node passes through the Softmax function to obtain a probability distribution $Q^{\ell}$ about the predicted class. In the following formula, $\phi^{\ell}$ is  the learned output of the $\ell \text{th}$ leaf node, thus we add $\phi$ to $\Theta$. The probability value of leaf node $\ell$ outputting class $k$ can be formulated as
\begin{equation}
\setlength{\abovedisplayskip}{5.5pt}
\setlength{\belowdisplayskip}{5.5pt}
Q_{k}^{\ell}=\frac{\exp \left(\phi_{k}^{\ell}\right)}{\sum_{k^{\prime}} \exp \left(\phi_{k^{\prime}}^{\ell}\right)}.
\end{equation}

In our paper, we use SDT as the base classifiers of boosting method. Specifically to the fraud detection task, a typical binary classification problem with two labels $\{-1, 1\}$, the output dimension of the leaf node can be $1$, meaning that a score function $h: \mathcal{X} \rightarrow \mathbb{R}$ is defined, and if $h(\boldsymbol{x}) \geqslant 0$, then the predicted label is 1, and otherwise -1.
\vspace{-2mm}
\subsubsection{Generating Soft Decision Tree}
\indent
Through the construction of leaf nodes and inner nodes above, each inner node generates left and right sub-paths, and each leaf node generates a probability distribution about the classes. Based on this, a hierarchical soft decision tree with the number of nodes $2^d -1$ can then be formed, where $d$ denotes the depth of the tree. A schematic diagram of SDT with depth 4 for binary classification can be found in Figure \ref{sdt}. The specific decision-making process includes:

\begin{figure}[t]
\centering
\includegraphics[height=3in]{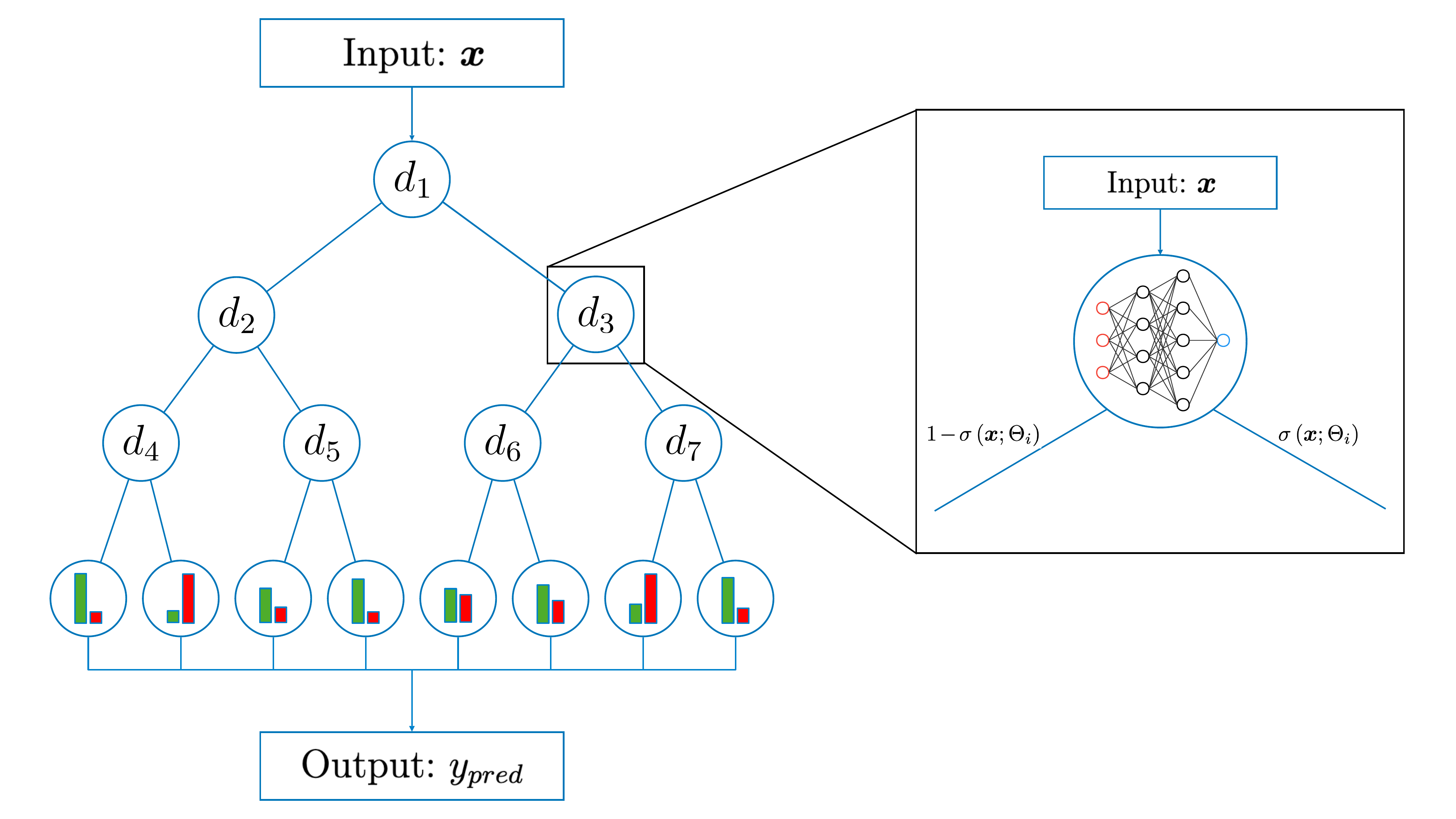}
\caption{Schematic diagram of SDT with each inner node is a multilayer neural network with depth of 4 for binary classification.} \label{sdt}
\end{figure}

\textbf{Path Probability:} Assuming that we choose a sample $\boldsymbol{x}$ as the model input, and our soft decision tree will split based on the previously mentioned selected sub-paths, which will form a specific decision-making path. Define 
$\pi_{i}(\boldsymbol{x} \mid \Theta)=\sum_{1 \leqslant j<i} d_{j}(\boldsymbol{x} ; \Theta_j)^{\mathbb{I}_{j}^{r}}\left(1-d_{j}(\boldsymbol{x} ; \Theta_j)\right)^{\mathbb{I}_{j}^{\ell}}$ 
as the probability that a input sample $\boldsymbol{x}$ reaches the leaf node $i$, where $d_{j}(\boldsymbol{x} ; \Theta_j)$ is the probability of choosing the right sub-path for the inner node $j$, and $\Theta=\{\Theta_j\}$ denotes the model parameters.

\textbf{Predicted Outcome:} Each leaf node will output a distribution about the class probability, and the final decision result of the soft decision tree is the weighted average $\mathbb{P}[y=k \mid \boldsymbol{x }, \Theta]=\sum_{\ell} Q_{k}^{\ell} \pi_{\ell}(\boldsymbol{x} \mid \Theta)$ by averaging the distributions over all the leaf nodes weighted by their respective path probabilities, where $Q_{k}^{ \ell}$ means the probability of the leaf node $\ell$ outputting  class $k$.
\vspace{-2mm}
\subsection{Integrating SDTs using Gradient Boosting}

In this section, we discuss the construction of deep boosting decision trees by integrating SDTs using gradient boosting, its objective function and optimization algorithm.
\vspace{-2mm}
\subsubsection{Constructing the Basic Structure Based on Gradient Boosting}
\indent
Given training data $\mathcal{T}=\{(\boldsymbol{x}_z, y_z)\}_{z=1}^N$, gradient boosting machine aims to approximate the real prediction function by an additive model $H(\boldsymbol{x})=\sum_{t=1}^{T}\beta_th_t(\boldsymbol{x})$, where $h_t(\cdot)$ is the base learner, and $\beta_t$ is the corresponding coefficient. In the training phase, $H(\boldsymbol{x})$ can be optimized by minimizing an empirical loss $\sum_{z=1}^Nl(H(\boldsymbol{x}_z),y_z)$ gradually with the more base learners as follows: based on the current model $H_m(\boldsymbol{x})=\sum_{t=1}^{m}\beta_th_t(\boldsymbol{x})$, computing a residual for each training sample $r_z^m=-{\partial l(H_m(\boldsymbol{x}_z),y_z)}/{\partial H_m(\boldsymbol{x}_z))}$, and then fitting the next base learner $h_{t+1}(\boldsymbol{x})$ towards the residuals, next coefficient $\beta_{t+1}$ can be determined by either least squares or a constant, and finally the model can be updated by $H_{m+1}(\boldsymbol{x})=H_{m}(\boldsymbol{x})+\beta_{t+1}h_{t+1}(\boldsymbol{x})$. We can see that gradient boosting machine uses the negative gradient as the residual and determines the base learners by fitting the residual in a sequential fashion. 
\newline
Obviously, gradient boosting is a serial structure meaning that the construction of the current base learner depends on the result of the previous one, which is a natural contradiction with the end-to-end parallel training framework of deep learning. To employ SDTs as base learner and integrate them to a neural network with gradient boosting like structure, we can first build $T$ SDTs $h_1(\boldsymbol{x},\Theta_1)$, $h_2(\boldsymbol{x},\Theta_2)$,..., $h_T(\boldsymbol{x},\Theta_T)$, where $\Theta_1$, $\Theta_2$,..., $\Theta_T$ are the parameters of corresponding SDTs. To
jointly optimize those SDTs, we construct local loss for each SDT as follows: $L_t = \sum_{z=1}^N[h_t(\boldsymbol{x}_z)-r_z^{t-1}]^2$, where $r_z^{t-1}$ denotes the residual calculated as above, and then those SDTs can be jointly parallel optimized by minimize a global loss $L = \sum_{t=1}^TL_t$. In this way, we can force each SDT to fit the residuals from the previous step, and finally obtain a end-to-end structure, which we called deep boosting decision trees (DBDT). Based on the trained model, for a new sample $\boldsymbol{x}$ we can make a prediction by $H(\boldsymbol{x})=\sum_{t=1}^{T}h_t(\boldsymbol{x})$.

\begin{figure}[t]
\centering
\includegraphics[width=\textwidth]{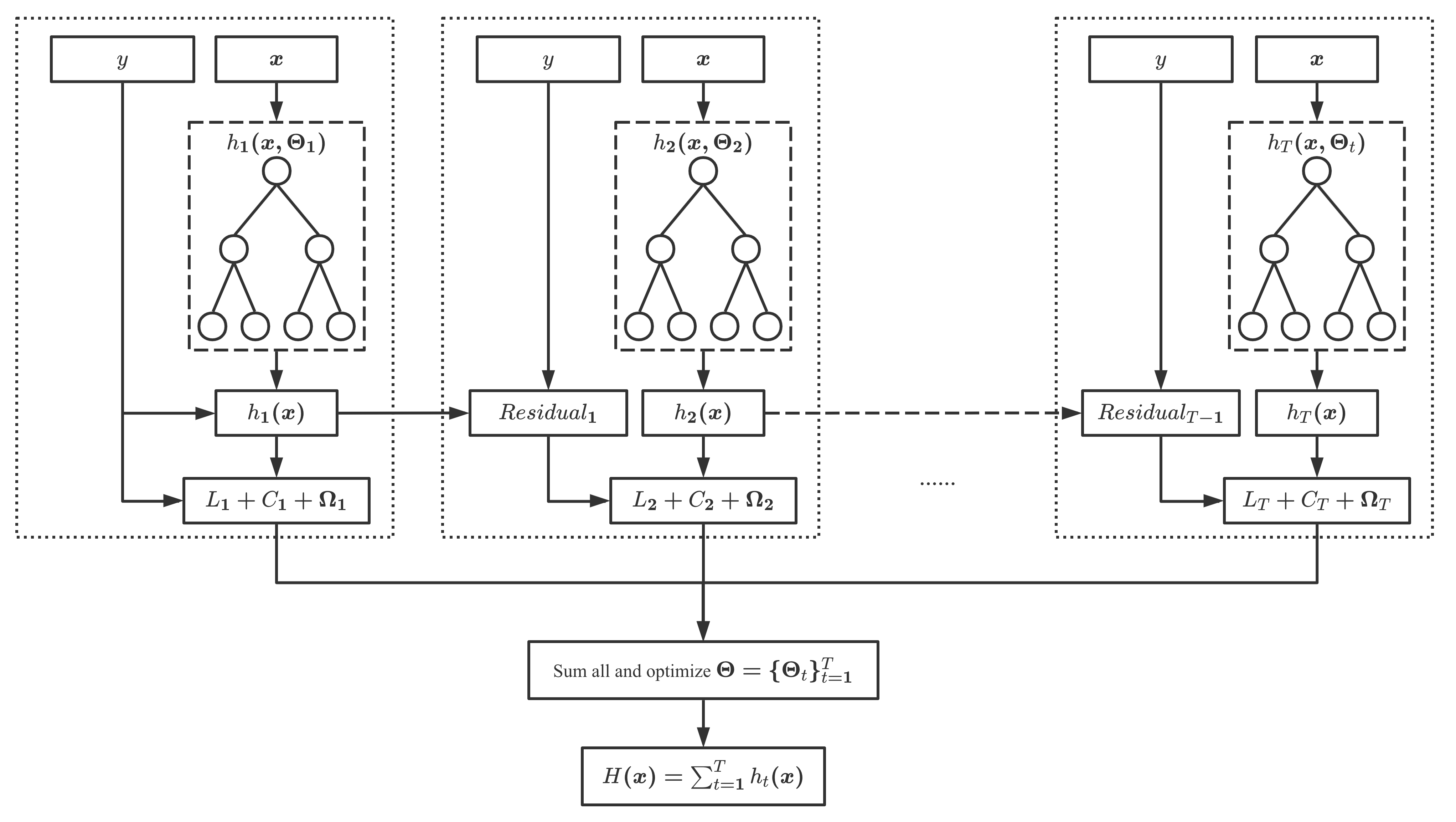}
\caption{The overall structure of DBDT.} \label{diagram}
\end{figure}
\vspace{-2mm}
\subsubsection{Objective Function}
\indent
In order to obtain the final model, it is necessary to construct a suitable objective loss function and get the parameters of our model by minimizing it. According to the conventional gradient boosting, we apply exponential loss function as the empirical loss, i.e., in the calculation of residuals we have $l\left(H\left(\boldsymbol{x}_{z}\right), y_{z}\right) = \exp{(-y_zH(\boldsymbol{x}_z))}$, and then residual can be computed by $r_z= y_z\exp{(-y_zH(\boldsymbol{x}_z))}$. To prevent the model from suffering from under- or over-fitting, we formulate an objective function that incorporates two regularization terms as: 
\begin{equation}\label{objective}
\setlength{\abovedisplayskip}{5.5pt}
\setlength{\belowdisplayskip}{5.5pt}
L_{Exp}=\sum_{t=1}^{T}\left(L_t+C_{t}+\Omega_{t}\right),
\end{equation}
where $C_t$ and $\Omega_{t} $ are the two regularization terms of the $t$th soft decision tree, aiming at controlling the importance of inner node and inner node feature importance, respectively.

\textbf{First Regularization Term} The first regularization term aims to mitigate the issue of under-fitting caused by insufficient diversity in sample data during training, which can lead to sub-nodes becoming overly reliant on specific sub-paths. To hedge against this, the expected  probability distribution of the selection of the two sub-paths should be $\{0.5, 0.5\}$, and the true distribution in a single-layer soft decision tree can be denoted as $\{\alpha_i, (1 - \alpha_i)\}$ with


\begin{equation}
\setlength{\abovedisplayskip}{5.5pt}
\setlength{\belowdisplayskip}{5.5pt}
\alpha_{i}=\frac{\sum_{\boldsymbol{x}} \pi_{i}(\boldsymbol{x} \mid \Theta) d_{i}(\boldsymbol{x} ; \Theta_i)}{\sum_{\boldsymbol{x}} \pi_{i}(\boldsymbol{x} \mid \Theta)},
\end{equation}

Thus we can use the cross entropy of the expected distribution $\{0.5, 0.5\}$ and the true distribution $\{\alpha_i, (1 - \alpha_i)\}$ as a penalty term to construct a SDT with uniform sub-path selection.
Considering that the number of the inputs falling on a certain node decays exponentially with the depth of the decision tree $d$, we add a decay coefficient $2^{-d}$ into this penalty and get the first regularization term of the $t$th soft decision tree as:
\begin{equation}
\setlength{\abovedisplayskip}{5.5pt}
\setlength{\belowdisplayskip}{5.5pt}
C_{t}=-\lambda_{1} \times 2^{-d} \sum_{i \in \text { Inner Nodes }} 0.5 \log \left(\alpha_{i}\right)+0.5 \log \left(1-\alpha_{i}\right),
\end{equation}
where $\lambda_1$ is a model hyperparameter.


\textbf{Second Regularization Term} The second regularization term aims to prevent overfitting of the model by imposing a penalty on the norm of the weight vector associated with inner nodes in the neural network. Therefore, the second regularization term of the $t$th SDT is
\begin{equation}
\setlength{\abovedisplayskip}{5.5pt}
\setlength{\belowdisplayskip}{3.5pt}
\Omega_{t}=\lambda_{2} \sum_{i \in \text { Inner Nodes }}|\Theta_i|_{2}
\end{equation}
where $\lambda_2$ is a model hyperparameter and $\Theta_i$ denotes the parameters of the neural network in the $i$th inner node.
\vspace{-2mm}
\subsubsection{Optimization Algorithm}
\indent
After construction, a DBDT can be written as $H(\boldsymbol{x};\Theta)=\sum_{t=1}^{T}h_t(\boldsymbol{x};\Theta_t)$, where $h_t(\boldsymbol{x};\Theta_t)$ is the SDT base learner with paremeters $\Theta_t$, and $\Theta=\{\Theta_t\}_{t=1}^T$ denotes the set of all parameters in DBDT.  DBDT can be seen as a neural network with gradient boosting decision trees like structure, and thus its parameters can be jointly optimized by back propagation (BP) algorithm and stochastic gradient descent (SGD) based on the objective function (\ref{objective}). We illustrate the overall structure of DBDT in Figure \ref{diagram}, and summary the optimization procedure using SGD in Algorithm \ref{alg1} which we call DBDT-SGD. 

\begin{algorithm}[!t]
	\renewcommand{\algorithmicrequire}{\textbf{Input:}}
	\renewcommand{\algorithmicensure}{\textbf{Return:}}
	\caption{DBDT-SGD}
	\label{alg1}
	\begin{algorithmic}[1]
	    \REQUIRE Training Set: $\mathcal{T}=\left\{\left(\boldsymbol{x}_{i}, y_{i}\right)\right\}_{i=1}^N$, $ \boldsymbol{x}_{i} \in \mathcal{X}=\mathbb{R}^{p}$, $y_{i} \in \mathcal{Y}=\{-1,+1\}$, number of SDTs: $T$, depth of SDT: $d$, layer number of inner nodes $c$, coefficients of regularization $\lambda_1$ and $\lambda_2$, number of epochs: $nEpochs$;
            \FOR {$t = 1, \ldots, T$} 
		      \STATE Construct SDT $h_t(\boldsymbol{x};\Theta_t)$ with depth $d$ and layer number $c$, and initialize parameters $\Theta_t$  using Xavier initialization;
            \ENDFOR
		\FOR {$i = 1, \ldots, nEpochs$}
            \STATE Let $h_{0}(\boldsymbol{x}) = 0$, $L_{Exp} = 0$, $H(\boldsymbol{x}) = 0$;  \\
            \FOR {$t = 1, \ldots, T$}
		        \STATE Compute the current SDT's output $h_t(\boldsymbol{x}_i;\Theta_t)$ for all $\left(\boldsymbol{x}_{i}, y_{i}\right) \in \mathcal{T}$; \\
		        \STATE Compute residuals $r_i= y_i\exp{(-y_iH(\boldsymbol{x}_i))}$ for all $\left(\boldsymbol{x}_{i}, y_{i}\right) \in \mathcal{T}$; \\
                \STATE Update $H(\boldsymbol{x}) = H(\boldsymbol{x})+h_t(\boldsymbol{x}_i;\Theta_t)$;
                \STATE Compute the local loss of $h_t(\boldsymbol{x})$: $L_t = \sum_{i=1}^N[h_t(\boldsymbol{x}_i)-r_i]^2$; \\
                \STATE Compute the two regularization terms of  $h_t(\boldsymbol{x})$: $C_{t}$,  $\Omega_{t}$;
                \STATE Update the objective function: $L_{Exp}=L_{Exp}+\left(L_t+C_{t}+\Omega_{t}\right)$;\ENDFOR
            \STATE Update $\Theta=\{\Theta_t\}_{t=1}^T$ w.r.t. $L_{Exp}$ using SGD;
            \ENDFOR
		\ENSURE $H(\boldsymbol{x})=\sum_{t=1}^{T}h_t(\boldsymbol{x})$.
	\end{algorithmic}  
\end{algorithm}
\vspace{-2mm}
\subsection{Dealing with imbalanced problems using DBDT}

DBDT-SGD has a significant effect on balanced data, but owing to fraud is a small probability event, fraud detection can translate into how to better deal with imbalanced data substantially. In imbalanced data, the minority class has little impact on the optimization of the model, leading to that the parameters of the model are almost determined by the majority class \cite{tantithamthavorn2018impact}. As a result, a surrogated loss based on accuracy and optimization algorithms such as SGD and Adam seriously degenerate in the imbalanced data. Stochastic AUC maximization method can be a possible way to deal with the imbalanced data prediction problems based on deep learning framework. Statistically, AUC (Area Under the ROC curve) is defined as the probability that the predicted score of positive samples is higher than the predicted score of negative samples \cite{hanley1982meaning}, \cite{clemenccon2008ranking}. Compared with accuracy and the corresponding surrogated loss, AUC is a more reasonable measure for imbalanced data, and meanwhile the AUC-based surrogate loss function also has better performance when training deep learning models with imbalanced data. To be specific, let $\mathbf{z}=(\boldsymbol{x}, y) \sim \mathbb{P}$ represents random data following an unknown distribution $\mathbb{P}$, where $\boldsymbol{x} \in \mathcal{X}$ is the feature vector, $y \in \mathcal{Y} = \{-1, 1 \}$ is its label, and $\mathcal{Z}=\mathcal{X} \times \mathcal{Y}$, $p=\operatorname{Pr}(y=1)=\mathbb{E}_{y}\left[\mathbb{ I}_{[y=1]}\right]$, where $\mathbb{I}(\cdot)$ denotes the indicator function. The AUC of a DBDT scoring model $H(\boldsymbol{x};\Theta)$ at the population level can be expressed as
\begin{equation}
\operatorname{AUC}(H)=\operatorname{Pr}\left(H(\boldsymbol{x};\Theta) \geq H\left(\boldsymbol{x}^{\prime};\Theta \right) \mid y=1, y^{\prime}=-1\right)
\end{equation}
where $\mathbf{z}=(\boldsymbol{x},y)$ and $\mathbf{z}^{\prime}=\left(\boldsymbol{x}^{\prime}, y^{\prime}\right)$ are independent of $\mathbb{P}$. We employ the $\ell_2$ loss as the surrogate function of the indicator function $\mathbb{I}(\cdot)$, and according to the nonlinear and non-convex characteristics of DBDT, the surrogated loss of AUC is defined as:
\begin{equation}
\setlength{\abovedisplayskip}{5.5pt}
\setlength{\belowdisplayskip}{5.5pt}
\min _{\Theta} P(\Theta):= \mathbb{E}_{\mathbf{z}, \mathbf{z}^{\prime}}\left[\left(1-H(\boldsymbol{x}; \Theta)+H\left(\boldsymbol{x}^{\prime}; \Theta \right)\right)^{2} \mid y=1, y^{\prime}=-1\right].
\end{equation}

The above formulates the surrogate loss of AUC as a non-convex function, that is, solving the surrogate loss of AUC is not a convex optimization problem. Thus, considering the duality principle, when the surrogate loss of AUC is difficult to solve, the AUC maximization problem can be transformed into a new dual problem using Lagrangian dual function. The benefit of such implement is that the new dual problem is convex and easy to solve. Then we make the following assumptions: (1) $\phi(\mathbf{v})$ satisfies PL condition; (2) $H(\mathbf{x}; \Theta)$ is $\tilde{L}- Lipschitz \ continuous$; (3) $\phi(\mathbf{v})$ is $L-smooth$; (4) $0 \leq H(\boldsymbol{x}; \Theta) \leq 1$; (5) $\operatorname{Var}[H(\boldsymbol{x}; \Theta) \mid y=-1] \leq \sigma^{2}$; (6) $\operatorname{Var}[H(\boldsymbol{x}; \Theta) \mid y=1] \leq \sigma^{2}$; (7) $\phi\left(\overline{\mathbf{v}}_{0}\right)-\phi\left(\mathbf{v}_{*}\right) \leq \Delta_{0}$, where $\mathbf{v}_{*}$ is the global minimum of $\phi$ and $ \Delta_{0} > 0 $. Under those assumptions, the AUC maximization problem can be finally transformed into
\begin{equation}
\setlength{\abovedisplayskip}{5.5pt}
\setlength{\belowdisplayskip}{5.5pt}
\begin{aligned}
& \min _{\Theta,(a, b) \in \mathbb{R}^{2}} \max _{\alpha \in \mathbb{R}} \Psi(\Theta, a, b, \alpha):=\mathbb{E}_{\mathbf{z}}[\psi(\Theta, a, b, \alpha ; \mathbf{z})], \\
\end{aligned}\label{auc max}
\end{equation}
where $\mathbf{z}=(\boldsymbol{x}, y) \sim \mathbb{P}$ and $\psi(\Theta, a, b, \alpha ; \mathbf{z})$ is defined as:
\begin{equation}
\setlength{\abovedisplayskip}{5.5pt}
\setlength{\belowdisplayskip}{5.5pt}
\begin{aligned}
&\psi(\Theta, a, b, \alpha ; \mathbf{z}) = (1-p)(H(\boldsymbol{x}; \Theta)-a)^{2} \mathbb{I}_{[y=1]} +p(H(\boldsymbol{x}; \Theta)-b)^{2} \mathbb{I}_{[y=-1]} \\
&~~~~~+2(1+\alpha)(p H(\boldsymbol{x}; \Theta) \mathbb{I}_{[y=-1]} -(1-p) H(\boldsymbol{x}; \Theta) \mathbb{I}_{[y=1]})-p(1-p) \alpha^{2},
\end{aligned}
\end{equation}
where $\Theta$ is the parameters to be trained of the DBDT, $a, \ b, \ \alpha$ is derived parameters to be trained for maximizing AUC.
Let $\mathbf{v}=\left(\Theta^{\top}, a, b\right)^{\top}$, $ \phi(\mathbf{v})=\max _{\alpha} \Psi(\mathbf{v}, \alpha)$, then it is obvious that for any $\mathbf{v}=\left(\Theta^{\top}, a, b\right)^{\top}$, we have $\min _{\Theta } P(\Theta)=\min _{\mathbf{v}} \phi(\mathbf{v})$ and $P(\Theta) \leq \phi(\mathbf{v})$. 
At this point, the loss function based on AUC maximization is proposed, and we map a non-convex optimization problem into a concave min-max stochastic optimization problem, where the original variable is convex to the loss function and the dual variable is concave to the loss function. Then we further define an  AUC loss function as:
\begin{equation}
\setlength{\abovedisplayskip}{5.5pt}
\setlength{\belowdisplayskip}{5.5pt}
L_{\mathrm{AUC}}(\Theta) :=\min _{(a, b) \in \mathbb{R}^{2}} \max _{\alpha \in \mathbb{R}} \Psi(\Theta, a, b, \alpha),
\end{equation}
then the min-max optimization problem \ref{auc max} can be transformed into $\min_\Theta L_{\mathrm{AUC}}(\Theta)$, which can then be effectively solved with theoretical convergence using a primal-dual algorithm Proximal Primal-Dual Stochastic Gradient\cite{liu2019stochastic}. 

\vspace{-2mm}
\subsection{Using Compositional Training for DBDT}

Training a deep neural network by maximizing the AUC leads to the degradation of the feature representation \cite{yuan2021compositional}, so training a DBDT by optimizing the AUC surrogated loss usually does not yield satisfactory performance. To avoid feature representation degradation, we optimize the DBDT through a compositional training approach for end-to-end deep AUC maximization\cite{yuan2021compositional}. Different from the AUC maximization, the key idea is to minimize a compositional objective function, where the external function of the objective function corresponds to an AUC loss, and the internal function represents an exponential loss function that minimizes the DBDT. The specific form of our compositional objective function is:
\begin{equation}
\setlength{\abovedisplayskip}{5.5pt}
\setlength{\belowdisplayskip}{5.5pt}
\min _{\Theta} L_{\mathrm{AUC}}\left(\Theta-\beta \nabla L_{\mathrm{AVG}}(\Theta)\right)
\end{equation}
where $\beta$ is a hyperparameter, $L_{\mathrm{AVG}}(\Theta) = \sum_{z=1}^NL_{\mathrm{Exp}}(\Theta;\boldsymbol{x}_{z},y_z)$ denotes the averaged exponential loss. We refer to the above objective as the compositional objective and a method for minimizing the above compositional objective as compositional training. 
We can see that $\Theta-\beta \nabla L_{\mathrm{AVG}}(\Theta)$ is computed by a gradient descent step for minimizing the averaged loss $L_{\mathrm{AVG}}(\Theta)$, which facilitates the learning of lower layers for feature extraction due to equal weights of all examples. In addition, taking a gradient descent step for the outer function $L_{AUC}(\cdot)$ will make the classifier more robust to the minority class due to the
higher weights of examples from the minority class. Overall, we have two alternating conceptual steps, i.e., the inner gradient descent step $\Theta-\beta \nabla L_{\mathrm{AVG}}(\Theta)$ acts as a feature purification step, and the outer gradient descent step
$\Theta-\eta\left(I-\beta \nabla^{2} L_{\mathrm{AVG}}(\Theta)\right) \nabla L_{\mathrm{AUC}}(\Theta-\beta \nabla L_{\mathrm{AVG}}(\Theta))$ acts as a classifier robustification step, where $\eta$ is a step size \cite{yuan2021compositional}. In particular, for the considered AUC loss, the compositional objective becomes:
\begin{equation}
\setlength{\abovedisplayskip}{5.5pt}
\setlength{\belowdisplayskip}{5.5pt}
\begin{aligned}
&\min _{\Theta,(a, b) \in \mathbb{R}^{2}} \max _{\alpha \in \mathbb{R}} \Psi\left(\Theta-\beta \nabla L_{\mathrm{AVG}}(\Theta), a, b, \alpha\right) \\  &=\frac{1}{N} \sum_{z=1}^{N} \psi\left(\Theta-\beta \nabla L_{\mathrm{AVG}}(\Theta), a, b, \alpha ; \boldsymbol{x}_{z}, y_{z}\right).
\end{aligned}
\end{equation}
\begin{algorithm}[!t]
	\renewcommand{\algorithmicrequire}{\textbf{Input:}}
	\renewcommand{\algorithmicensure}{\textbf{Return:}}
	\caption{Primal-Dual Stochastic Compositional Adaptive (PDSCA)}
	\label{alg3}
	\begin{algorithmic}[1]
	    \STATE  Require Parameters: $\beta_{0}$, $\beta_{1}$, $\beta$, $G_{0}$, $ \eta_{1}$, $ \eta_{2}$; number of epochs: $nEpochs$;
		\STATE Initialization: $\overline{\Theta}_{0}=\left(\Theta_{0}; a_{0}; b_{0}\right) \in \mathbb{R}^{d+2}$, $\alpha_{0}$, $\mathbf{u}_{0} \in \mathbb{R}^{d+2}$;
		\FOR {$i = 1, \ldots, nEpochs$}
            \STATE Sample two set of example denoted by $\mathcal{S}_1$, $\mathcal{S}_2$;  \\
            \STATE Use a moving average technique to purify feature: $\mathbf{u}_{i+1}=\left(1-\beta_{0}\right) \mathbf{u}_{i}+\beta_{0} q\left(\overline{\Theta}_{i} ; \mathcal{S}_{1}\right)$; \\
            \STATE Estimate the gradient of the outer function:  \\
            $\mathcal{O}_{i}=\nabla_{\overline{\Theta}} q\left(\overline{\Theta}_{i} ; \mathcal{S}_{1}\right)^{\top}\left[\nabla_{\mathbf{u}} g_{1}\left(\mathbf{u}_{i+1} ; \mathcal{S}_{2}\right)+\alpha_{t} \nabla_{\mathbf{u}} g_{2}\left(\mathbf{u}_{i+1} ; \mathcal{S}_{2}\right)\right]$;  \\
            \STATE Use momentum methods for updating $\overline{\Theta}$: $\mathbf{z}_{i+1}=\left(1-\beta_{1}\right) \mathbf{z}_{i}+\beta_{1} \mathcal{O}_{i}$;   \\
            \STATE Compute the adaptive step size: $\mathbf{z}_{2, i+1}=q_{i}\left(\left\{\mathcal{O}_{j}, j=0, \ldots, i\right)\right\}$; \\
            \STATE Update the $\overline{\Theta}$: $\overline{\Theta}_{i+1}=\overline{\Theta}_{i}-\eta_{1} \frac{\mathbf{z}_{i+1}}{\sqrt{\mathbf{z}_{2, i+1}}+G_{0}}$;   \\
            \STATE Update the dual variable: \\ $\alpha_{i+1}=\Pi_{\mathbb{R}}\left[\alpha_{i}+\eta_{2}\left(g_{2}\left(\mathbf{u}_{i+1} ; \mathcal{S}_{1} \cup \mathcal{S}_{2}\right)-\nabla g_{3}\left(\alpha_{i}\right)\right)\right]$;  \\
        
        \ENDFOR
		\ENSURE $\overline{\Theta}, \alpha$.
	\end{algorithmic}  
\end{algorithm}

Denote $\overline{\Theta}=(\Theta ; a ; b)$ for simplify of presentation, and we can write $\psi\left(\overline{\Theta}, \alpha ; \boldsymbol{x}_{z}, y_{z}\right)$ as:
\begin{equation}
\setlength{\abovedisplayskip}{5.5pt}
\setlength{\belowdisplayskip}{5.5pt}
\psi\left(\overline{\Theta}, \alpha ; \boldsymbol{x}_{z}, y_{z}\right)=g_{1}\left(\overline{\Theta} ; \boldsymbol{x}_{z}, y_{z}\right)+\alpha g_{2}\left(\overline{\Theta} ; \boldsymbol{x}_{z}, y_{z}\right)-g_{3}(\alpha),
\end{equation}
where
\begin{equation}
\setlength{\abovedisplayskip}{5.5pt}
\setlength{\belowdisplayskip}{5.5pt}
\begin{aligned}
&g_{1}\left(\overline{\Theta} ; \boldsymbol{x}_{z}, y_{z}\right)=(1-p)\left(H\left(\boldsymbol{x}_{z} ; \Theta \right)-a\right)^{2} \mathbb{I}_{\left[y_{z}=1\right]} +p\left(H\left(\boldsymbol{x}_{z} ; \Theta\right)-b\right)^{2} \mathbb{I}_{\left[y_{z}=-1\right]} \\
&~~~~~~~~~~~~~~~~~~~+2 p H\left(\boldsymbol{x}_{z} ; \Theta\right) \mathbb{I}_{\left[y_{z}=-1\right]}-2(1-p) H\left(\boldsymbol{x}_{z} ; \Theta\right) \mathbb{I}_{\left[y_{z}=1\right]}; \\
&g_{2}\left(\overline{\Theta} ; \boldsymbol{x}_{z}, y_{z}\right)=2 p H\left(\boldsymbol{x}_{z} ; \Theta\right) \mathbb{I}_{\left[y_{z}=-1\right]}-2(1-p) H\left(\boldsymbol{x}_{z} ; \Theta\right)\mathbb{I}_{\left[y_{z}=1\right]}; \\
&g_{3}(\alpha)=p(1-p) \alpha^{2}.
\end{aligned}
\end{equation}

For the sake of brevity, we sample a set of data denoted by $\mathcal{S}$, and $g_{1}(\overline{\Theta} ; \mathcal{S})=\frac{1}{|\mathcal{S}|} \sum_{z \in \mathcal{S}} g_{1}\left(\overline{\Theta} ; \boldsymbol{x}_{z}, y_{z}\right)$, $g_{2}(\overline{\Theta} ; \mathcal{S})=\frac{1}{|\mathcal{S}|} \sum_{z \in \mathcal{S}} g_{2}\left(\overline{\Theta} ; \boldsymbol{x}_{z}, y_{z}\right)$, and define $q(\overline{\Theta})$, $\nabla_{\overline{\Theta}} q(\overline{\Theta})$, $q(\overline{\Theta} ; \mathcal{S})$ as follows:
\begin{equation}
\setlength{\abovedisplayskip}{5.5pt}
\setlength{\belowdisplayskip}{5.5pt}
\begin{aligned}
&q(\overline{\Theta})=\left(\Theta-\beta \nabla L_{\mathrm{AVG}}(\Theta) ; a ; b\right); \\
&\nabla_{\overline{\Theta}} q(\overline{\Theta})=\left(I-\beta \nabla_{\Theta}^{2} L_{\mathrm{AVG}}(\Theta) ; 1 ; 1\right); \\
&q(\overline{\Theta} ; \mathcal{S})=\left(\Theta-\beta \nabla L_{\mathrm{AVG}}(\Theta ; \mathcal{S}) ; a ; b\right).
\end{aligned}
\end{equation}

Then the compositional AUC maximization approach can be summarized in Algorithm \ref{alg3}, which we call Primal-Dual Stochastic Compositional Adaptive (PDSCA).
\end{spacing}
\vspace{-4mm} 
\section{Benchmarks}
\begin{spacing}{2.0}
\vspace{-2mm} 
We employ some state-of-the-art methods, including conventional and deep learning models, as baselines to highlight the effectiveness of the proposed DBDT. Those compared methods and corresponding hyperparameters settings are described in detail below.

\begin{itemize}
\item[$\bullet$]
Random Forest (RF): Random Forest is an ensemble learning method that is widely used in classification and regression problems \cite{ho1998random}. According to the default hyperparameters of sklearn.RandomForestClassifier, we set the depth of decision tree as 4, and the number of decision trees as 100.

\item[$\bullet$]
Gaussian Naive Bayes (NB): Gaussian Naive Bayes is a probabilistic classification algorithm based on Bayes theorem with strong independence assumptions \cite{john2013estimating}. In line with the default of sklearn.GaussianNB, We set the portion of the largest variance of all features as $1e^{-9}$.

\item[$\bullet$] 
AdaBoost: AdaBoost is one of the most commonly-used boosting algorithms, and it has proven to be effective and easy to implement in various classification problems. According to the default of sklearn.AdaBoostClassifier, we set the depth of tree as 4, the number of weak classifiers 100 and learning ratio 0.1.

\item[$\bullet$]
XGBoost: XGBoost is an optimized distributed gradient boosting algorithm designed to be highly efficient, flexible and portable. On the basis of the hyperparameter settings of other boosting algorithms, we set the depth of tree as 4, the number of weak classifiers 100 and learning ratio 0.1.

\item[$\bullet$]
LightGBM: LightGBM is a gradient boosting framework with decision tree as the base learner. We set the depth of tree as 4, the number of weak classifiers 100 and learning ratio 0.1.

\item[$\bullet$]
SMOTBoost: A hybrid method that uses AdaBoost.M2 and SMOTE method (a commonly used over-sampling approach for data balancing) to detect fraud events. We set the depth of tree as 4, the number of weak classifiers 100 and learning ratio 0.1.

\item[$\bullet$]
RUSBoost: A hybrid method that uses AdaBoost.M2 and random under-sampling methods to detect fraud events. We set the depth of tree as 4, the number of weak classifiers 100 and learning ratio 0.1.

\item[$\bullet$]
Self-Paced Ensemble: Self-Paced Ensemble uses an under-sampling + ensemble strategy for boosting-like serial training, and gets an additive model to detect fraud events. We set the number of estimators as 100 and base estimator as decision tree with depth 4. 

\item[$\bullet$]
Multilayer Perceptron (MLP): a fully connected deep neural network \cite{rosenblatt1961principles}. We totally set four hidden layers, and each hidden layer consists of 8642 nodes.

\item[$\bullet$]
MLP-AUC: we set MLP-AUC the same structure as the MLP above, and train the model using the proposed compositional AUC maximization approach.

\end{itemize}
\end{spacing}
\vspace{-6mm}
\section{Experimental design}
\begin{spacing}{2.0}
\vspace{-2mm}
\subsection{Datasets}
We use five real-life imbalanced datasets(Bank Marketing\footnotemark[1] \footnotetext[1]{\href{https://archive.ics.uci.edu/ml/datasets/bank+marketing}{https://archive.ics.uci.edu/ml/datasets/bank+marketing}}, Vehicle Insurance \footnotemark[2] \footnotetext[2]{\href{https://www.kaggle.com/datasets/shivamb/vehicle-claim-fraud-detection}{https://www.kaggle.com/datasets/shivamb/vehicle-claim-fraud-detection}}, Fraudulent on Cars \footnotemark[3] \footnotetext[3]{\href{https://www.kaggle.com/datasets/surekharamireddy/fraudulent-claim-on-cars-physical-damage}{https://www.kaggle.com/datasets/surekharamireddy/fraudulent-claim-on-cars-physical-damage}}, Worldline \& ULB\footnotemark[4] \footnotetext[4]{\href{https://www.kaggle.com/datasets/mlg-ulb/creditcardfraud}{https://www.kaggle.com/datasets/mlg-ulb/creditcardfraud}}, BankSim \footnotemark[5] \footnotetext[5]{\href{https://www.kaggle.com/datasets/ealaxi/banksim1}{https://www.kaggle.com/datasets/ealaxi/banksim1}}) of fraud detection to perform our analysis. Table \ref{table1} shows the summary statistics of those datasets. The targets are divided into two types: normal and abnormal, among them, majority of the data are normal, and minority are abnormal. The features of datasets can be divided into two categories: categorical features and numerical feature, and meanwhile categorical features include ordered and unordered features. 
\begin{table}[h!]
	\centering  
	\caption{Summary statistics of the five datasets}  
	\label{table1}  
	\begin{tabular}{ccccc}
        \toprule
        Dataset & Samples & Features & Ratio of Minority \\
        \midrule
        Bank Marketing & 36,168 & 17 & 11.61\% \\
        Vehicle Insurance & 11,564 & 32 & 5.98\% \\
        Fraudulent on Cars & 17,998 & 26 & 15.65\% \\
        Worldline \& ULB & 284,807 & 29 & 0.17\% \\
        BankSim & 594,643 & 5 & 1.21\% \\
        \bottomrule
    \end{tabular}
\end{table}

The datasets contain category features, and some of them are disordered among different categories. For instance, there is marital status of users in Bank Marketing. Since computers can only accept numeric inputs, the original category features need to be digitally encoded. In this paper we adopt  One-Hot Encoding to process this class of category features without sequential information. Specifically, the category features are mapped to a One-Hot vector. Taking gender as an example, this feature contains two categories, each corresponding to a one-hot vector of length 1, namely male = $\{1\}$ and female = $\{0\}$.

Similarly, the datasets also contain some sequential features among categories, such as Educational Background which contains 5 categories, including Lower secondary, Secondary/secondary special, Incomplete  higher, Higher education and Academic degree. For this kind of features, One-hot Encoding cannot be adopted because of the sequential information between different classes. Sequential Encoding is adopted in this paper, that is, the sequential features are mapped to different natural numbers. Taking Educational Background as an example, they are mapped to Lower secondary = $\{1\}$, Secondary/secondary special = $\{2\}$, Incomplete higher = $\{3\}$, Higher education = $\{4\}$, and Academic degree = $\{5\}$.

In addition to category features, there are also many numerical features in these datasets, such as the age of users in the Bank Marketing. When we input these numerical features to the model directly, it will cause some troubles due to the difference in the magnitude of different attributes. Therefore, we normalize the numerical features in this paper. 
\vspace{-2mm}
\subsection{Evaluation metrics}
We employ ROC curve and the corresponding AUC as the main evaluation metric to evaluate the performance of the compared approaches for fraud detection. 
First, we introduce the confusion matrix,
wich counts the number of samples with True Positive (TP) meaning actual class is positive and predicted class is also positive, False Positive (FP) predicted class is positive but actual class is negative, False Negative (FN) predicted class is negative but actual class is positive, and True Negative (TN) predicted class is negative and actual class is also negative. 

Based on confusion matrix, ROC curve and AUC can be calculated as follows. In classification task, the model usually outputs a prediction probability. We can sort these probability in descending order and select a truncation in the middle, then the front part is predicted as positive examples, and the rest negative examples. This method is too subjective for the selection of the threshold, so researchers introduced the ROC curve from the medical field. The ROC curve is obtained by plotting for each possible threshold value false positive rate ($FPR = FP / (TN + FP$)) on the horizontal axis, true positive rate ($TPR = TP / (TP + FN)$) on the longitudinal axis. 
In order to facilitate the comparison between different imbalance learning algorithms, researchers define the AUC to represent the Area Under the ROC Curve.  AUC acts as  a single score varing between 0 and 1, and in the context of fraud detection, the AUC of a classifier can be interpreted as the probability that a randomly chosen fraud case is predicted a higher score than a randomly chosen legitimate case. Therefore, a higher AUC indicates superior classification performance.

In addition to ROC curve and AUC, we further introduce another common metric for imbalanced classification, H‐measure, to assist a comprehensive evaluation. The H‐measure is a classifier performance measure which takes into account the context of application without requiring a rigid value of relative misclassification costs to be set\cite{japkowicz2013assessment}. For more details on the H-measure, reader can refer to \cite{hand2009measuring}. As AUC,  higher H‐measure indicates superior classification performance.
\end{spacing}
\vspace{-4mm}
\section{Results}
\begin{spacing}{2.0}
\vspace{-2mm}
\subsection{Fraud Detection}

In this section, we compare the performances of our method DBDT and those benchmarks on the aforementioned fraud detection datasets. The Pytorch library was used to implement DBDT. We use SGD and PDSCA to optimize the DBDT parameters, respectively, and name the resulting model DBDT-SGD and DBDT-Com. The training is performed for 200 epochs and $batch \ size = 128$. The DBDT hyperparameters are selected according to the default of popular tree boosting API, such as AdaBoost and XGBoost. The values finally selected are $T = 40$, $d = 4$, $\lambda_1 = 0.1$, $\lambda_2 = 0.005$. The layer number of neural network in inner node is $c = 1$. The hyperparameters of compositional deep AUC maximization are initialized as $margin = 1.0$, $\gamma = 500$, $T_0 = 200$, $K = 1000$, $weight dacay = e^{-4}$, $\beta_1 = 0.9$, $\beta_2 = 0.999$. The model estimation is performed on a Nvidia RTX 2080 Ti GPU server with 256GB of RAM.

\begin{figure}[h!]
\centering
\begin{minipage}[t]{0.32\textwidth}
\centering
\includegraphics[width=\textwidth]{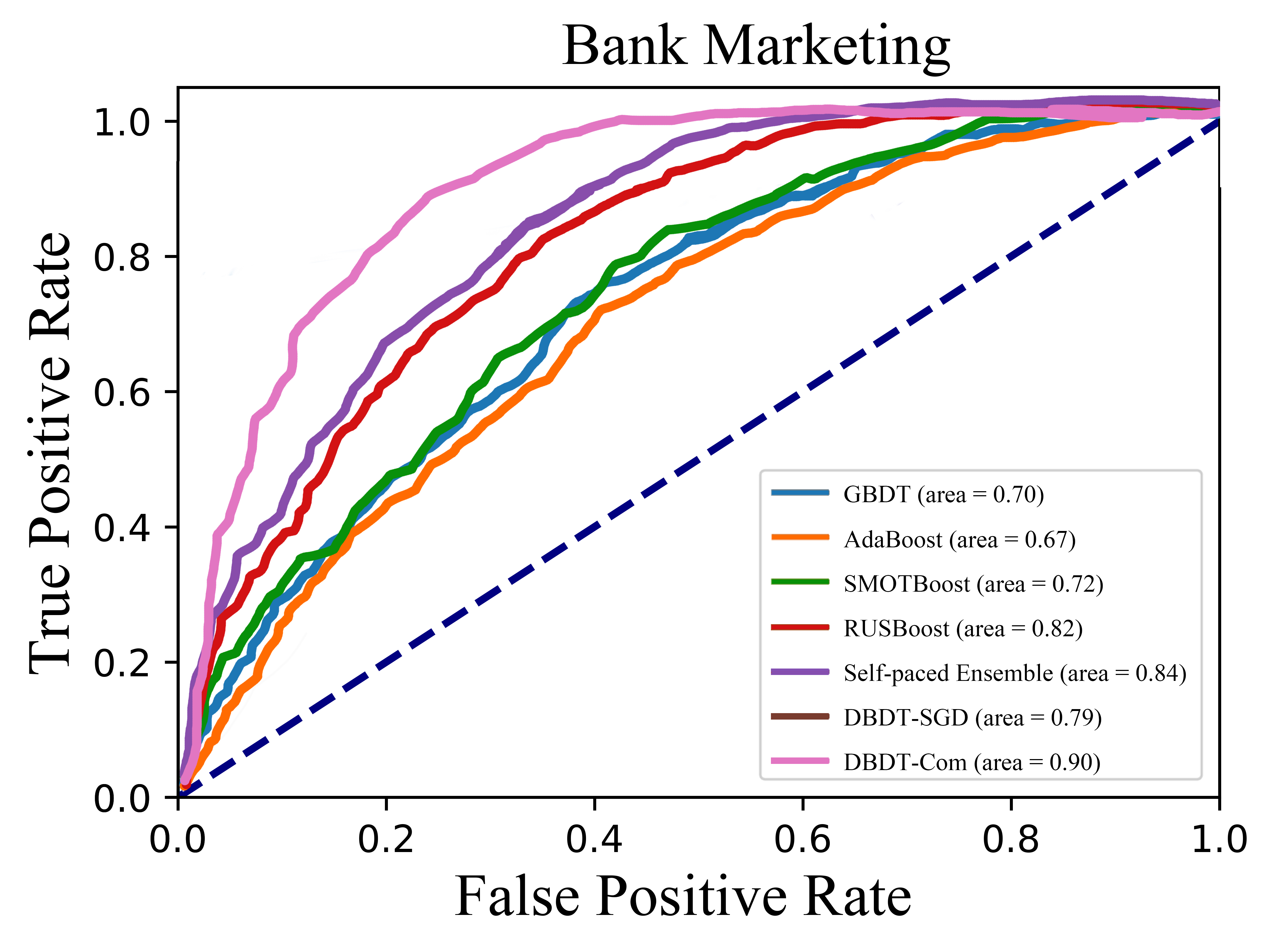}
\end{minipage}
\begin{minipage}[t]{0.32\textwidth}
\centering
\includegraphics[width=\textwidth]{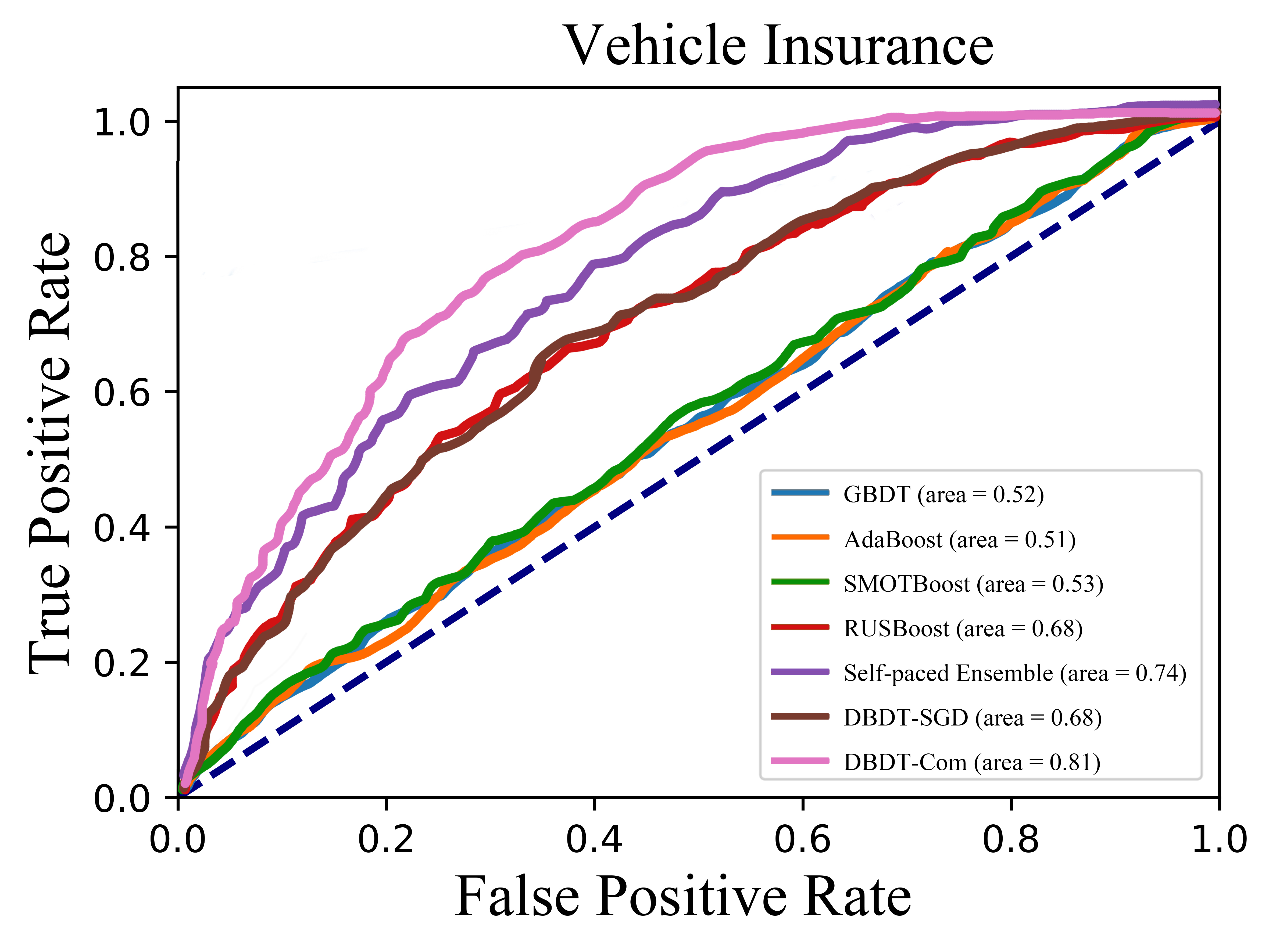}
\end{minipage}
\begin{minipage}[t]{0.32\textwidth}
\centering
\includegraphics[width=\textwidth]{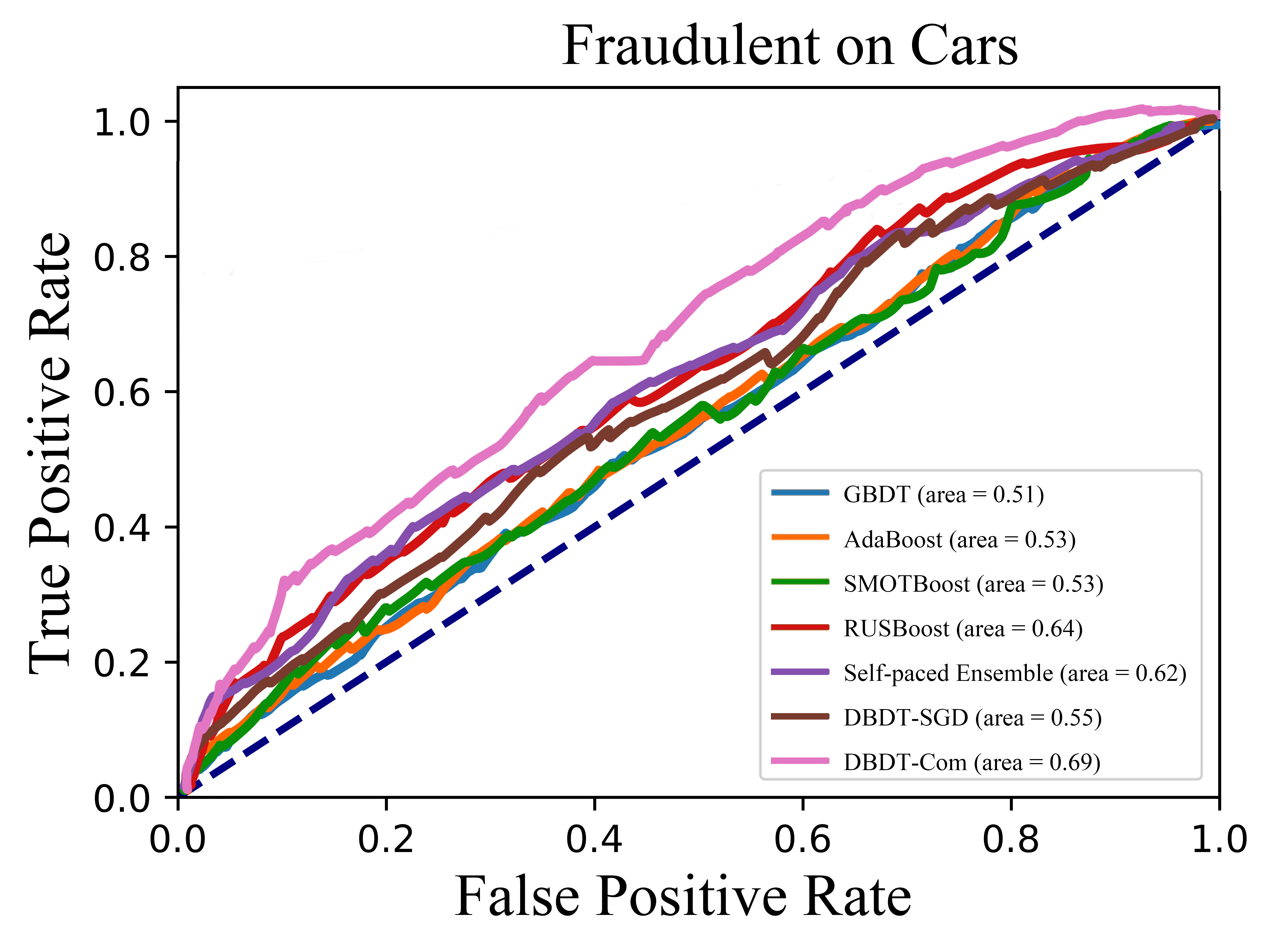}
\end{minipage}
\\
\begin{minipage}[t]{0.32\textwidth}
\centering
\includegraphics[width=\textwidth]{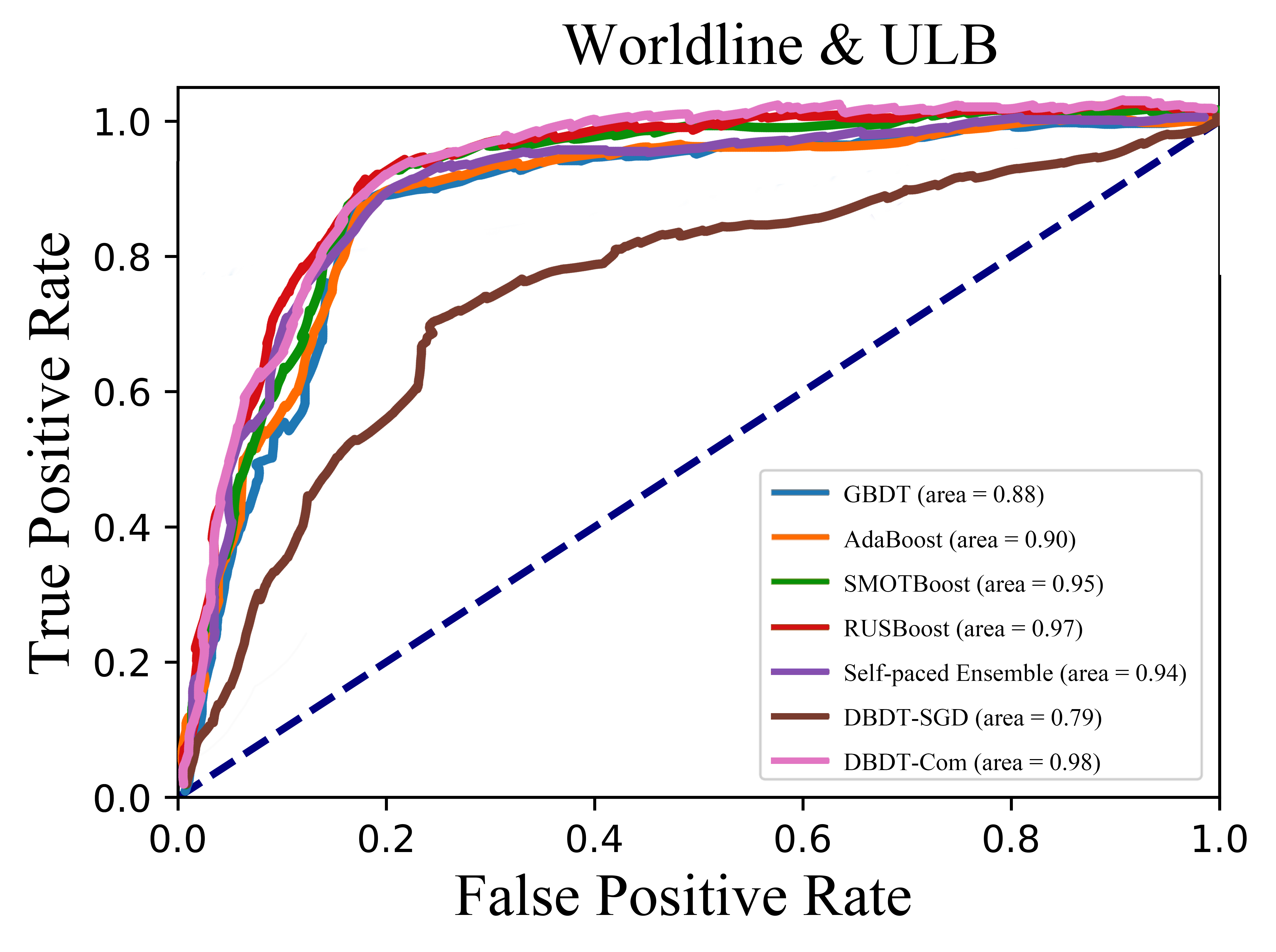}
\end{minipage}
\begin{minipage}[t]{0.32\textwidth}
\centering
\includegraphics[width=\textwidth]{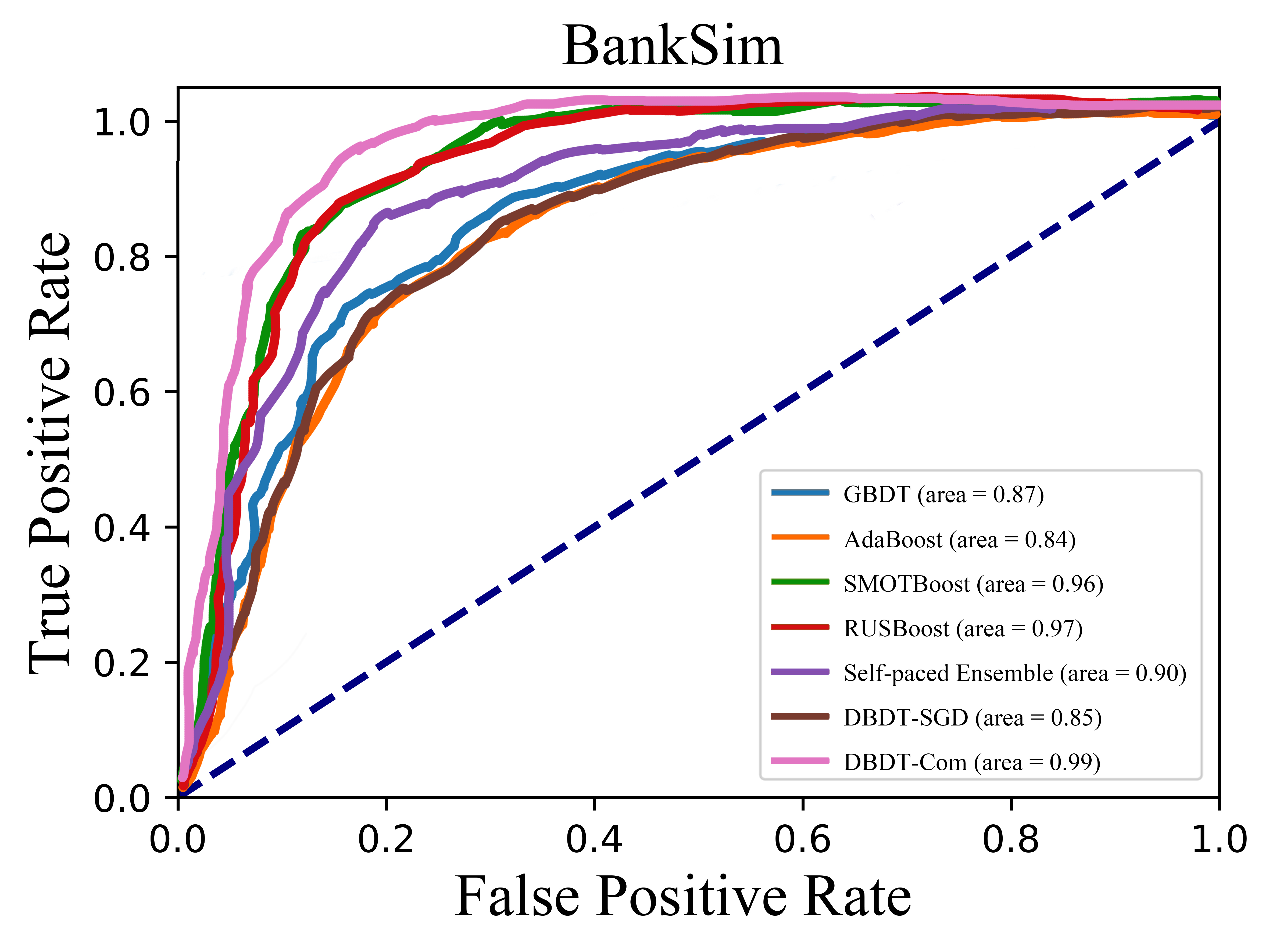}
\end{minipage}
\caption{The ROC curves of those compared methods on the five datasets.}\label{roc}
\end{figure}

We evaluate the performance of twelve models for the fraud detection, including all baseline mentioned above
and our method DBDT. Furthermore, to verify the effectiveness of our compositional AUC maximizaiton approach in DBDT for imbalanced data, we evaluate two DBDT models with different optimization algorithms, i.e., DBDT-SGD for algorithm \ref{alg1} which tries to minimize accuracy, and DBDT-Com for algorithm \ref{alg3} which tries to compositionally maximize AUC. For every dataset, we use 10-fold cross-validation for the model evaluation. The experimental results of means and standard deviations of AUC and H-measure are summarized in Table \ref{table2}, and meanwhile the ROC curves are plotted in Figure \ref{roc}.

\begin{table}[h!]
	\centering  
	\footnotesize
	\caption{Results of the fraud detection experiment}  
	\label{table2}  
	\begin{tabular}{cccccc}
        \toprule
        Dataset & Algorithm & AUC  &  $\sigma_{AUC}$  & H-measure  &  $\sigma_{H-measure}$ \\
        \midrule
        \multirow{12}{*}{\makecell[c]{Bank \\ Marketing}	}&RF&0.697362&0.009187&0.307367&0.017361 \\
        &NB&0.698348&0.006949&0.242479&0.014377 \\
        &MLP&0.725668&0.040664&0.377202&0.026382 \\
        &AdaBoost&0.671573&0.006154&0.259806&0.012630 \\
        &XGBoost&0.702117&0.005945&0.317574&0.013138 \\
        &LightGBM&0.660096&0.005042&0.246249&0.010758 \\
        &SMOTBoost&0.728006&0.012651&0.330803&0.026874 \\
        &RUSBoost&0.825999&0.008244&0.462583&0.019180 \\
        &SP-Ensemble&0.805472&0.009893&0.481405&0.020834 \\
        &MLP-AUC&0.715766&0.025345&0.365292&0.017825 \\
        &DBDT-SGD&0.807435&0.008267&0.382962&0.023903 \\
        &DBDT-Com&\textbf{0.908771}&0.007196&\textbf{0.519362}&0.019752 \\
        \midrule
        \multirow{12}{*}{\makecell[c]{Vehicle \\ Insurance}	}&RF&0.512952&0.007480&0.119542&0.011501 \\
        &NB&0.648543&0.021176&0.185174&0.021090 \\
        &MLP&0.575825&0.079460&0.147291&0.024482 \\
        &AdaBoost&0.508504&0.008685&0.110225&0.011344 \\
        &XGBoost&0.524166&0.013961&0.138397&0.022863 \\
        &LightGBM&0.518442&0.012542&0.129501&0.020364 \\
        &SMOTBoost&0.601733&0.021986&0.184487&0.030941 \\
        &RUSBoost&0.700925&0.057599&0.261147&0.057847 \\
        &SP-Ensemble&0.668632&0.027079&0.250737&0.042383 \\
        &MLP-AUC&0.619562&0.095725&0.238463&0.039252 \\
        &DBDT-SGD&0.680491&0.038626&0.253962&0.025936 \\
        &DBDT-Com&\textbf{0.812363}&0.015902&\textbf{0.279268}&0.029583 \\
        \midrule
        \multirow{12}{*}{\makecell[c]{Fraudulent \\ on Cars}	}&RF&0.503756&0.002706&0.105097&0.003703 \\
        &NB&0.515141&0.006209&0.113248&0.008929 \\
        &MLP&0.530672&0.039528&0.127636&0.016392 \\
        &AdaBoost&0.530307&0.006239&0.131704&0.008125 \\
        &XGBoost&0.518527&0.007727&0.118930&0.010526 \\
        &LightGBM&0.505278&0.002456&0.106601&0.003601 \\
        &SMOTBoost&0.531916&0.009428&0.125267&0.011594 \\
        &RUSBoost&0.657124&0.015361&0.206150&0.018800 \\
        &SP-Ensemble&0.606966&0.018625&0.157389&0.017337 \\
        &MLP-AUC&0.559736&0.026435&0.139425&0.017385 \\
        &DBDT-SGD&0.579262&0.018352&0.169258&0.013502 \\
        &DBDT-Com&\textbf{0.692563}&0.009386&\textbf{0.269374}&0.012948 \\
        \midrule
        \multirow{12}{*}{\makecell[c]{\makecell[c]{Worldline \\ \& ULB}}}&RF&0.896171&0.026373&0.769845&0.057971 \\
        &NB&0.904205&0.021602&0.762800&0.047452 \\
        &MLP&0.753045&0.063739&0.619735&0.083962 \\
        &AdaBoost&0.880683&0.031311&0.736061&0.068050 \\
        &XGBoost&0.896011&0.025537&0.769478&0.055684 \\
        &LightGBM&0.820394&0.072348&0.608868&0.145176 \\
        &SMOTBoost&0.941864&0.020578&\textbf{0.847416}&0.045055 \\
        &RUSBoost&0.914205&0.022079&0.748517&0.052447 \\
        &SP-Ensemble&0.917559&0.024827&0.816713&0.054546 \\
        &MLP-AUC&0.883053&0.073623&0.723960&0.062962 \\
        &DBDT-SGD&0.800125&0.039743&0.749206&0.059262 \\
        &DBDT-Com&\textbf{0.983615}&0.023085&0.833095&0.049285 \\
        \midrule
        \multirow{12}{*}{BankSim}&RF&0.875361&0.011197&0.720599&0.024413 \\
        &NB&0.884320&0.008752&0.728119&0.019364 \\
        &MLP&0.835025&0.026833&0.629378&0.025364 \\
        &AdaBoost&0.840807&0.009686&0.647376&0.020764 \\
        &XGBoost&0.873748&0.010924&0.718630&0.023842 \\
        &LightGBM&0.852320&0.011313&0.672069&0.024381 \\
        &SMOTBoost&0.951239&0.005811&0.861595&0.012918 \\
        &RUSBoost&0.873759&0.067207&0.670798&0.152472 \\
        &SP-Ensemble&0.886118&0.013042&0.743892&0.028426 \\
        &MLP-AUC&0.905736&0.058262&0.802852&0.048720 \\
        &DBDT-SGD&0.860319&0.019420&0.783956&0.029672 \\
        &DBDT-Com&\textbf{0.985467}&0.025315&\textbf{0.932058}&0.019582 \\
        \bottomrule
    \end{tabular}
\end{table}

From Table \ref{table2} and Figure \ref{roc}, we can see that on those fraud detection datasets, those general methods, Random Forest, Navie Bayes, MLP, AdaBoost, XGBoost and LightGBM, get poor performance, and actually there is a clear gap between these methods compared with others. This is because that they are originally designed for the general machine learning tasks, and fail to take into account the particularity of the fraud detection problem. They directly use raw data for the model training without any sampling strategy, and meanwhile use the accuracy as the optimization target, which leads to their poor adaptability to imbalanced data. SMOTBoost, RUSBoost and Self-Paced Ensemble try to alleviate the data imbalance issue through various re-sampling strategy. Table \ref{table2} and Figure \ref{roc} show that these re-sampling based methods can significantly improve the performance in those fraud detection datasets.  Specifically, Self-Paced Ensemble is a popular algorithm more recently for imbalanced learning, based on which we can verify whether our
method has advantages compared to the current state-of-the-arts.

As we said above, DBDT-SGD optimizes DBDT model using algorithm \ref{alg1}, meaning that it is a general machine learning algorithm without considering the data imbalance, which is consistent with the compared general classification methods. While we can observe from Table \ref{table2} that in most of those datasets DBDT-SGD significantly outperforms those general methods, which indicates that DBDT can act as a general machine learning algorithm with superior performance. Furthermore, MLP-AUC greatly outperforms MLP, and DBDT-Com achieves significantly higher performance than those compared methods including DBDT-SGD, from which we can conclude that: (1) our DBDT model with compositional AUC maximization strategy can achieve superior performance to state-of-the-arts on fraud detection, which shows the advantage of our approach; (2) the proposed compositional AUC maximization strategy can  effectively improve the adaptability of the algorithm to data imbalance, and thus has the potential to be extended to other algorithms.

\begin{figure}[h!] 
\centering
\includegraphics[width=\textwidth]{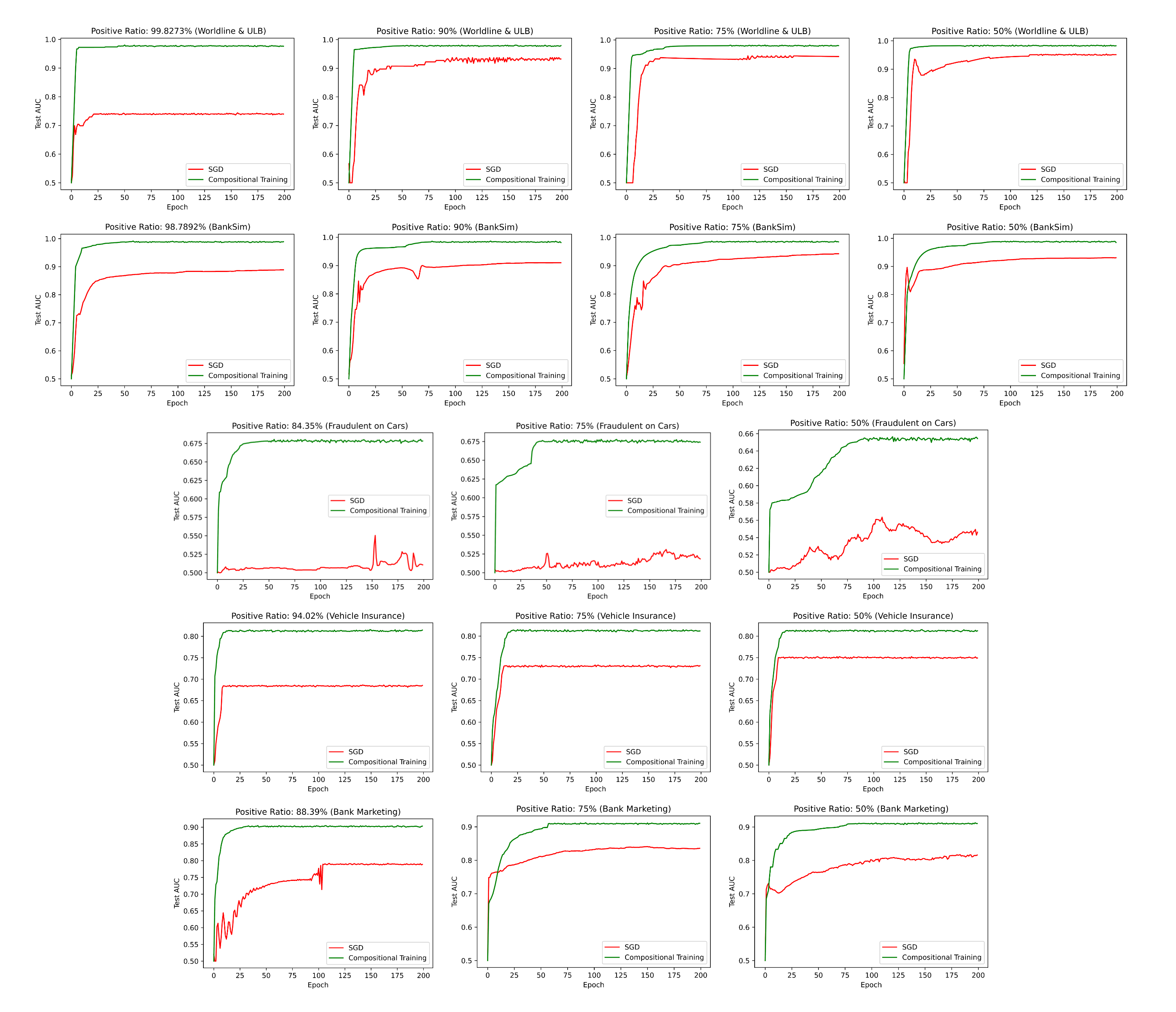}
\caption{Comparison of test AUC of SGD and Compositional Training based DBDT on fraud datasets with different positive ratios.}\label{auc}
\end{figure}

Next, we experimentally observe the influence of the positive ratio (ratio of the majority class) in fraud detection data over the performance of DBDT. To this end, we adopt a under-sampling strategy on the five fraud datasets respectively as follows: we keep all minority instances and random sample instances from all majority instances, and generate synthetic datasets with corresponding different positive ratios(50\%, 75\%, 90\% as well as original imbalanced ratio). Here, we need to notice that the original positive ratios of Fraudulent on Cars, Vehicle Insurance and Bank Marketing datasets are below or just above 90\%, thus, we do not set the 90\% positive ratio experiment on them. Then we train DBDT using SGD and Compositional Training strategy on those synthetic datasets, respectively. We report the results in Figure \ref{auc}, where each row is a different dataset, each column is a different positive ratio, and in each subfigure X-axis represents the training epochs, Y-axis test AUC.  We can see from Figure \ref{auc} that for the balanced settings with ratio equal to 50\%, the performance of SGD is consistently the closest to the Compositional Training, although always worse. And for imbalanced settings, Compositional Training are more advantageous than SGD in all cases and performs almost the same with various positive ratios, which shows that the Compositional Training strategy has good adaptability and robustness to data imbalance. In addition, we can observe that, as the training epoch grows, Compositional Training has better convergence than SGD.
\begin{figure}[h!]
\centering
\includegraphics[height=2in]{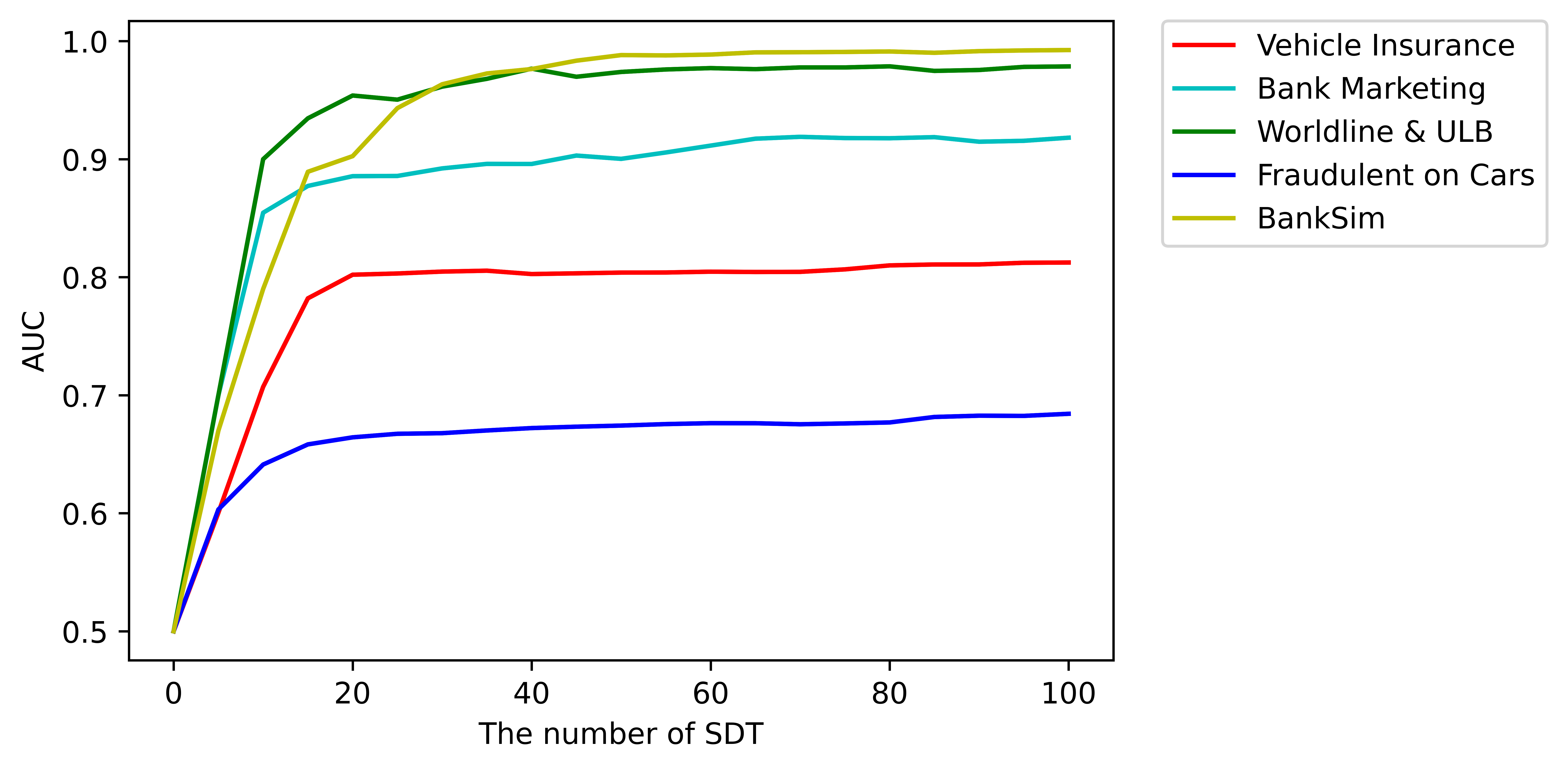}
\caption{The effect of the number of SDTs in DBDT.} \label{number}
\end{figure}

Finally, we employ a common statistical test approach, the Friedman test, to verify whether the performance of DBDT-Com is significantly different from the compared methods. First, the compared algorithms are sorted according to the corresponding value of AUC  or H-measure, where the algorithm with the best performance is assigned to 1, the second is assigned to 2, and so on. Let the null hypothesis be that there is no significant difference in the performance of these algorithms, and then the statistic of Friedman test can be computed by
\begin{equation}\label{Friedman test}
\tau_{F}=\frac{(N-1) \tau_{\mathcal{X}^{2}}}{N(k-1)-\tau_{\mathcal{X}^{2}}},
\end{equation}
where $N$ is the number of datasets, $k$ is the number of algorithms, and $\tau_{\mathcal{X}^{2}}$ is a random variable from $\mathcal{X}^{2}$ distribution. Random variable $\tau_{F}$ is from F-distribution, whose DOF is $k-1$ and $(k-1)(N-1)$. Substituting the sorting results into (\ref{Friedman test}), we can easily get $\tau_{F} = 38.230769$ for AUC and  $\tau_{F} = 33.061538$ for H-measure, which corresponds to the p value $7.152060e^{-5}$ and $0.000514$, respectively. We can see that the p values of AUC and H-measure are much smaller than the critical p value 0.05, and thus the null hypothesis is rejected at the significance levels of 0.05, which supports our claim that DBDT-Com significantly outperforms the compared methods  from a statistical test point of view.

\vspace{-2mm}
\subsection{Parameter Sensitivity}
In this section, we study the model sensitivity to hyperparameters, which includes the number of SDTs for tree boosting, the depth of SDT and the depth of neural network in each inner node. To explore the impact of these hyperparameters on model performance, we take the five fraud detection datasets as examples, use 5-fold cross validation, and average the metrics. The basic idea of ensemble learning is to integrate several weak classifiers to construct a strong classifier, so the number of weak classifiers is a very important hyperparameter. In order to explore the impact of the number of SDTs on the performance of DBDT, we build DBDT models with tree depth 4, number of neural network layers 1, penalty coefficients 0.1 and 0.005, and the number of SDTs varies from 5 to 100. We report the corresponding AUC changes in Figure \ref{number}, we can see that increasing the number of SDTs will improve the performance of the model, but there is an upper limit. As can be seen from Figure \ref{number}, when the number of SDTs increases to about 40, the model basically reaches the upper limit of performance and the improvement will be small.

\begin{figure}[h!]
\centering
\begin{minipage}[t]{0.32\textwidth}
\centering
\includegraphics[width=\textwidth]{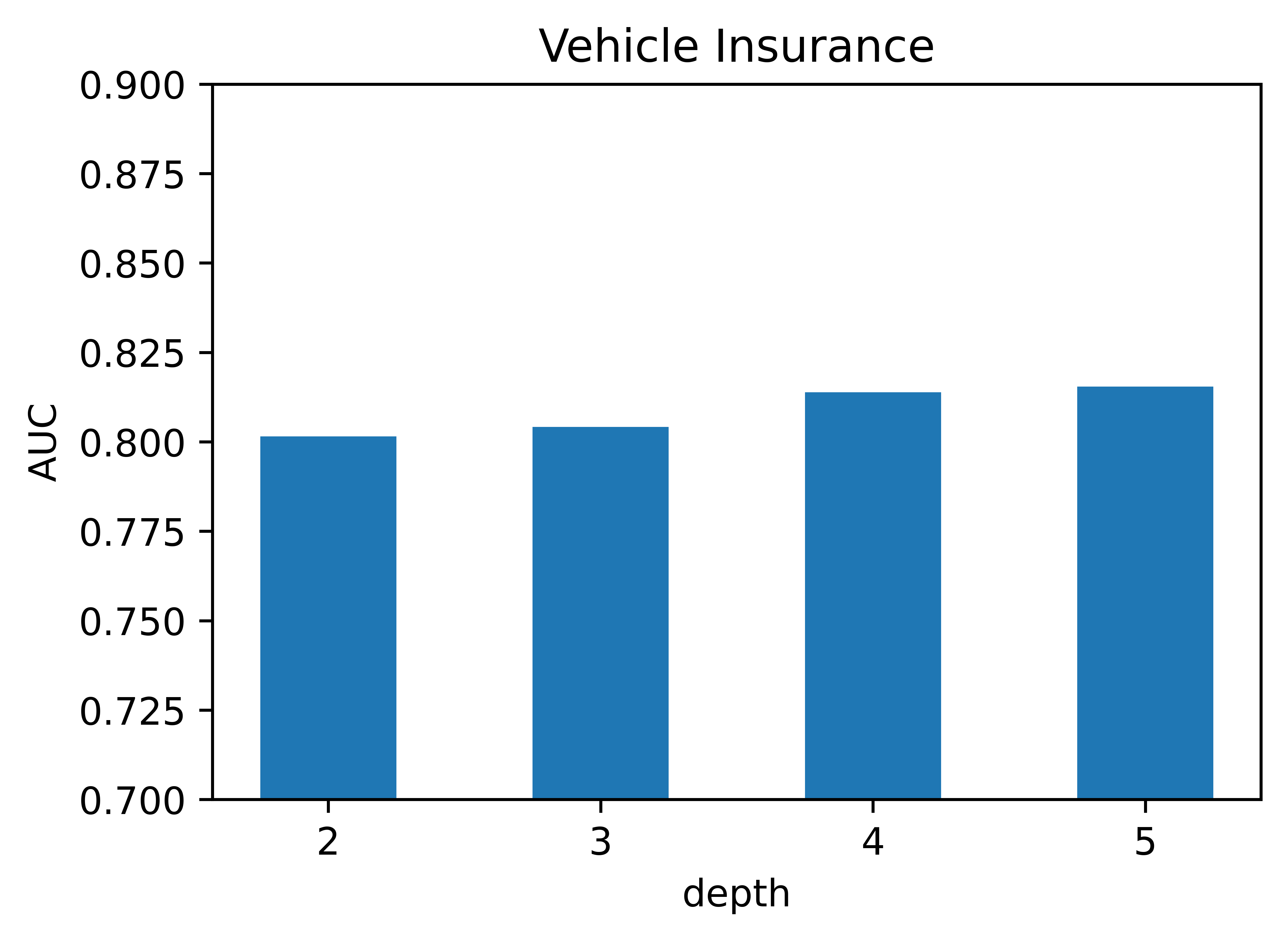}
\end{minipage}
\begin{minipage}[t]{0.32\textwidth}
\centering
\includegraphics[width=\textwidth]{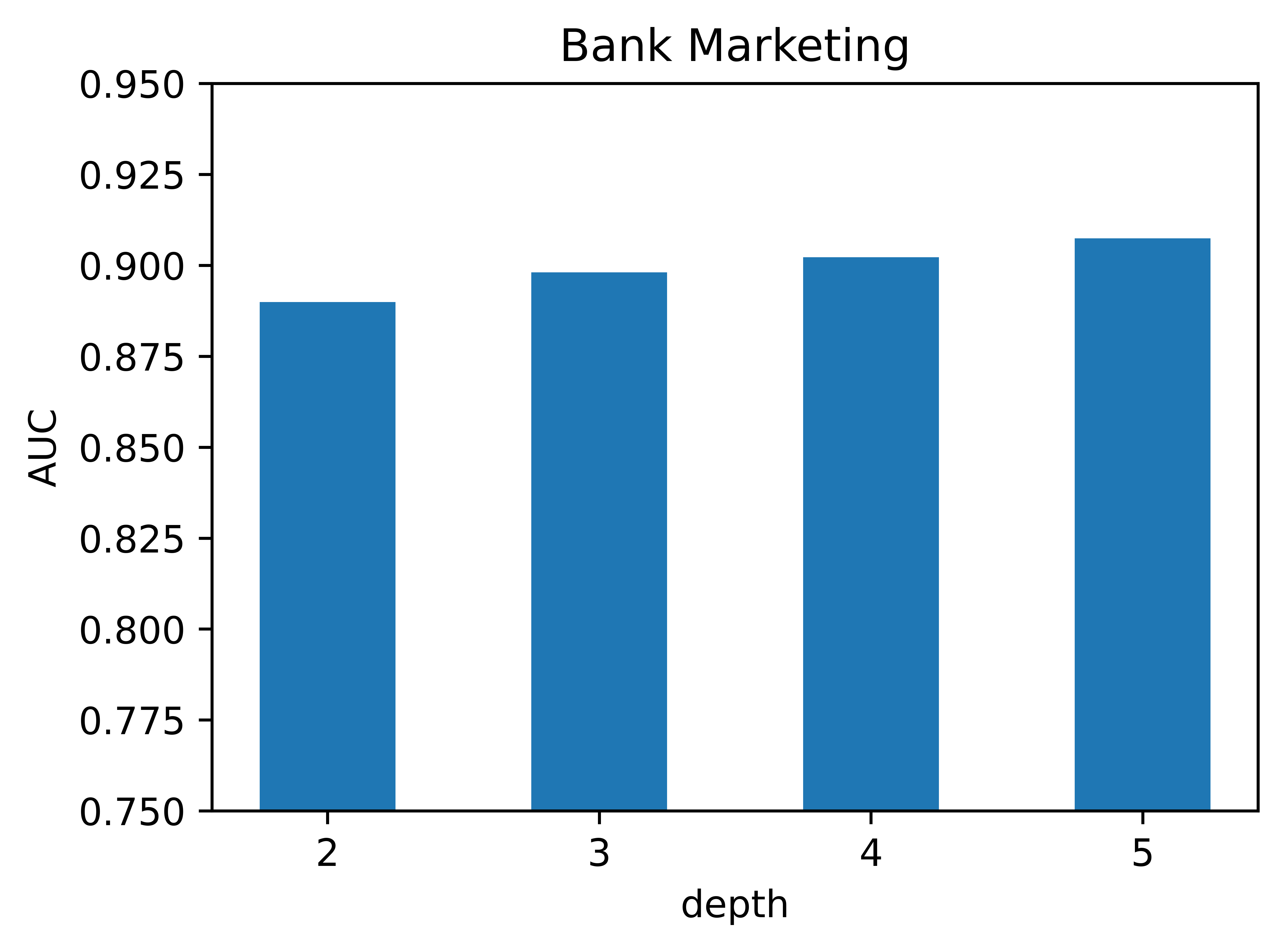}
\end{minipage}
\begin{minipage}[t]{0.32\textwidth}
\centering
\includegraphics[width=\textwidth]{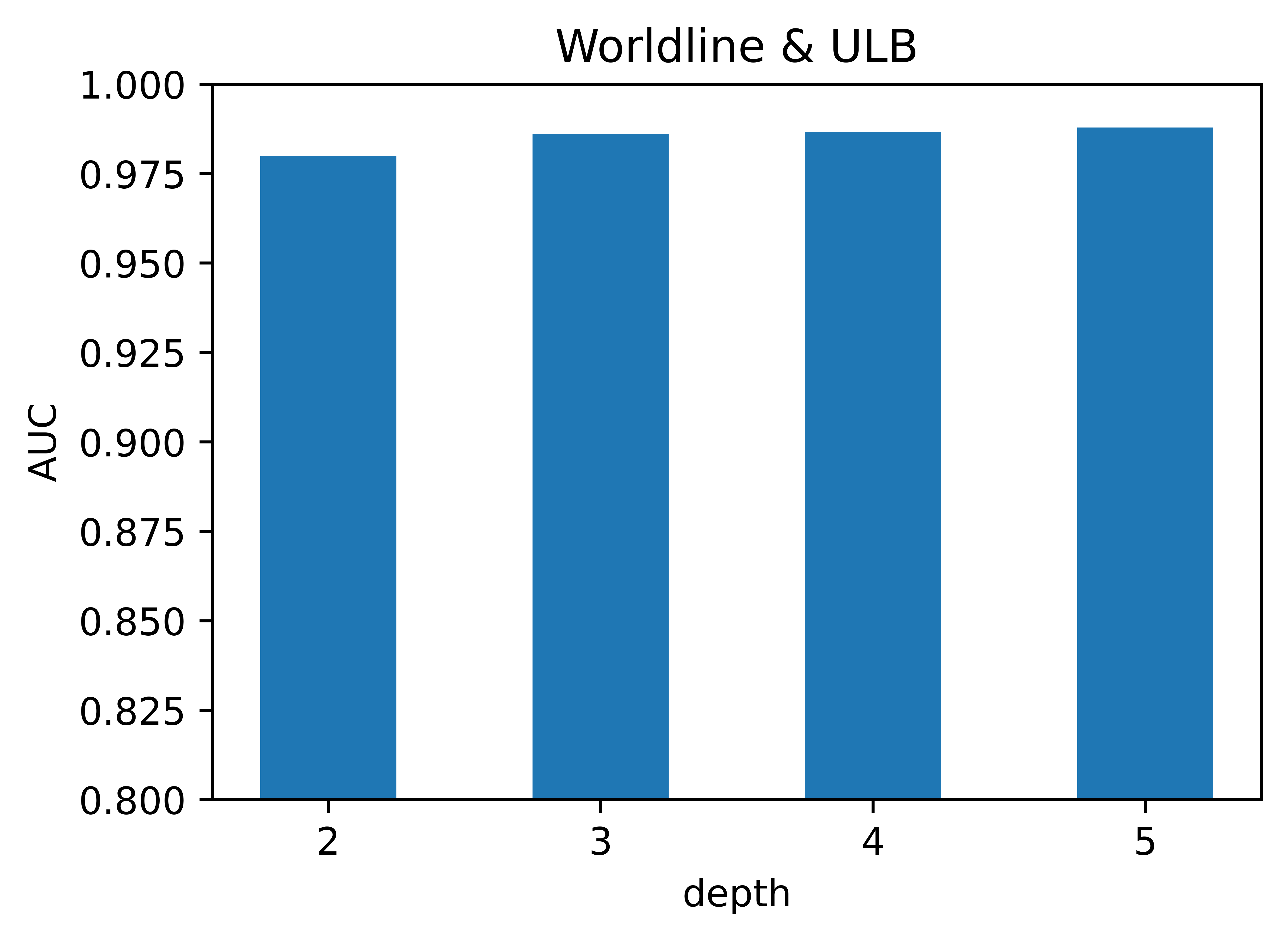}
\end{minipage}
\\
\begin{minipage}[t]{0.32\textwidth}
\centering
\includegraphics[width=\textwidth]{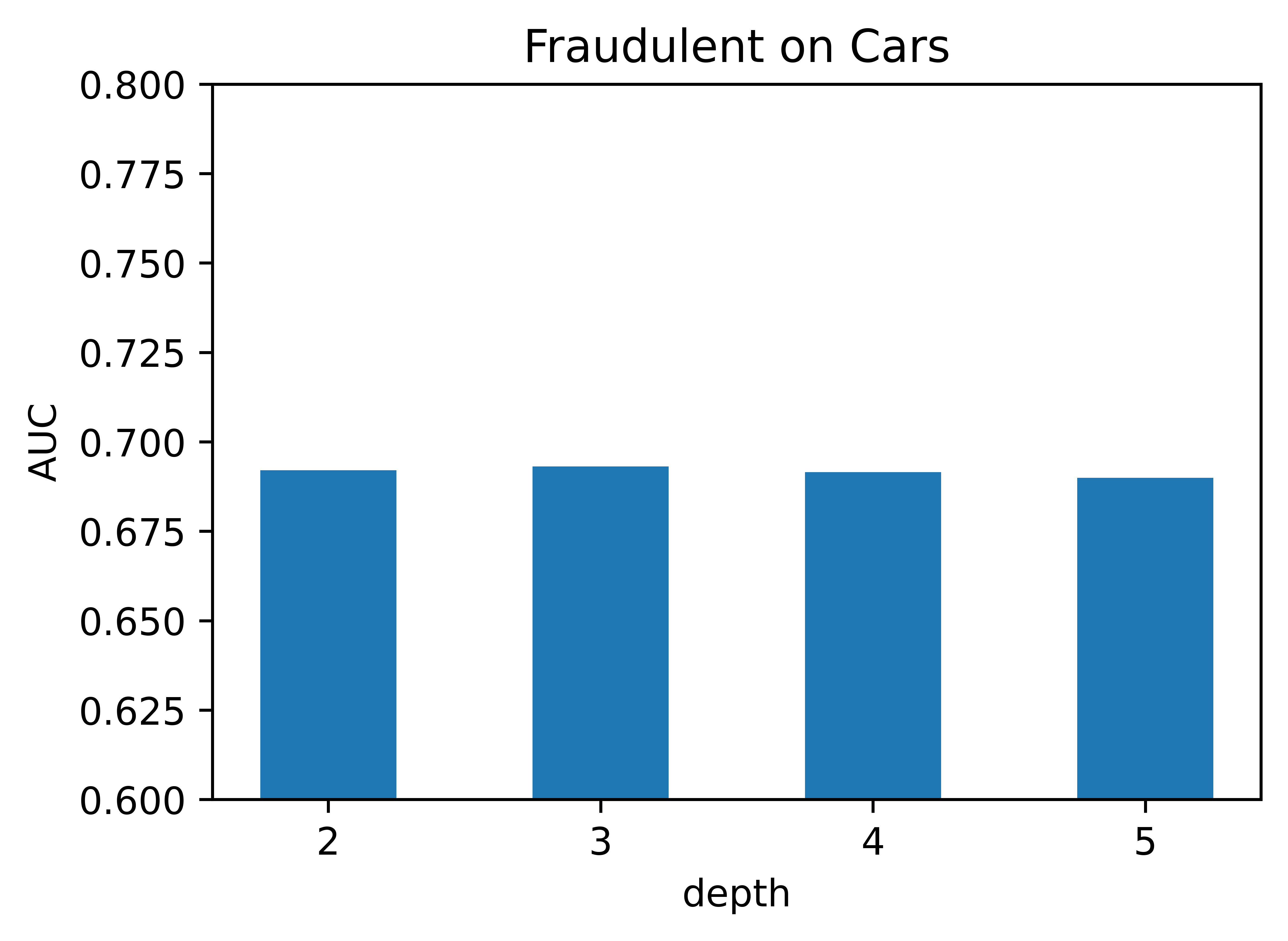}
\end{minipage}
\begin{minipage}[t]{0.32\textwidth}
\centering
\includegraphics[width=\textwidth]{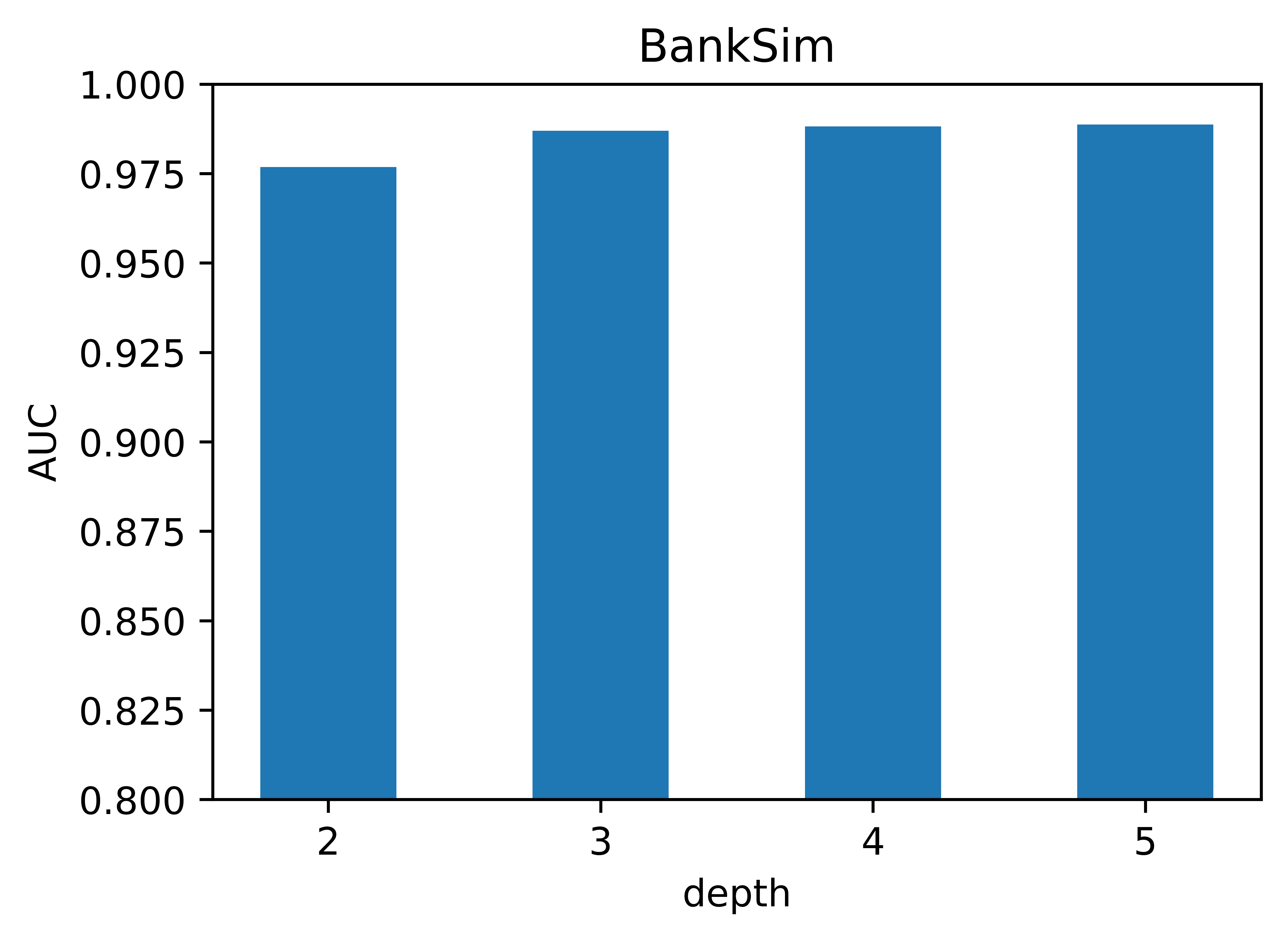}
\end{minipage}
\caption{The effect of depth of the SDTs in DBDT.}\label{depth}
\end{figure}

In addition to the number of SDTs, the depth of the SDT is also an important hyperparameter of DBDT. As mentioned above, the depth of SDT $d$ meets the exponential relationship $n = 2^d - 1$ with the number of nodes $n$ in the tree, and each node corresponds to a neural network. If the $d$ we select is too large, the complexity of the model will be greater than fraud detection requirements, so it is important to determine an appropriate tree depth. We fix other hyperparameters as: the number of SDTs is 40, the number of hidden layers of neural network is 1, and the regularization term coefficients $\lambda_1$ and $\lambda_2$ are 0.1 and 0.005. We change the depth of SDTs from 2 to 5 and observe the AUC changes. The results are reported in Figure \ref{depth}. We can see that increasing the depth of the SDT improves model performance, but there is an upper limit which can be reached when the depth is up to 4 and 5. 

\begin{figure}[h!]
\centering
\begin{minipage}[t]{0.32\textwidth}
\centering
\includegraphics[width=\textwidth]{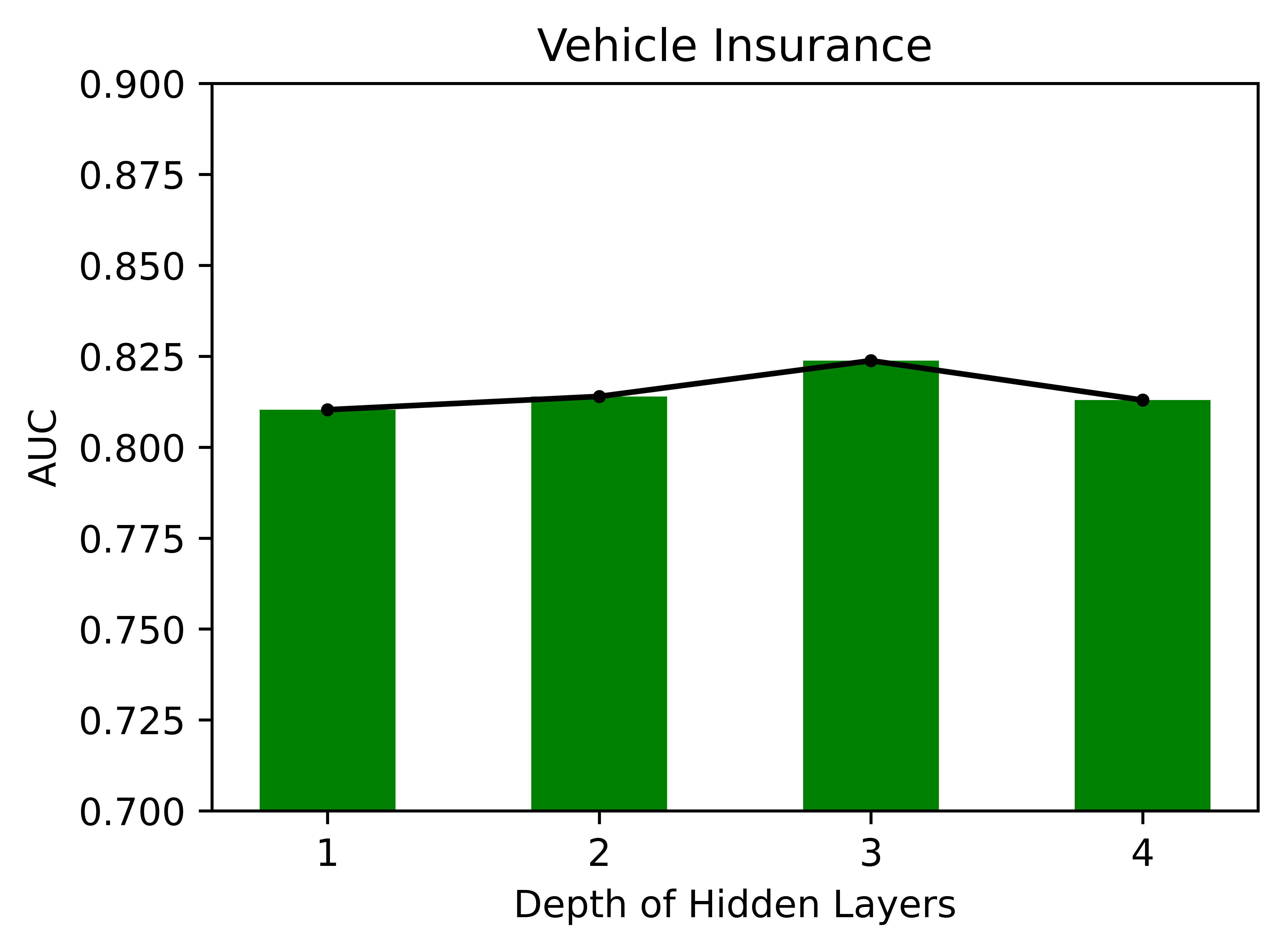}
\end{minipage}
\begin{minipage}[t]{0.32\textwidth}
\centering
\includegraphics[width=\textwidth]{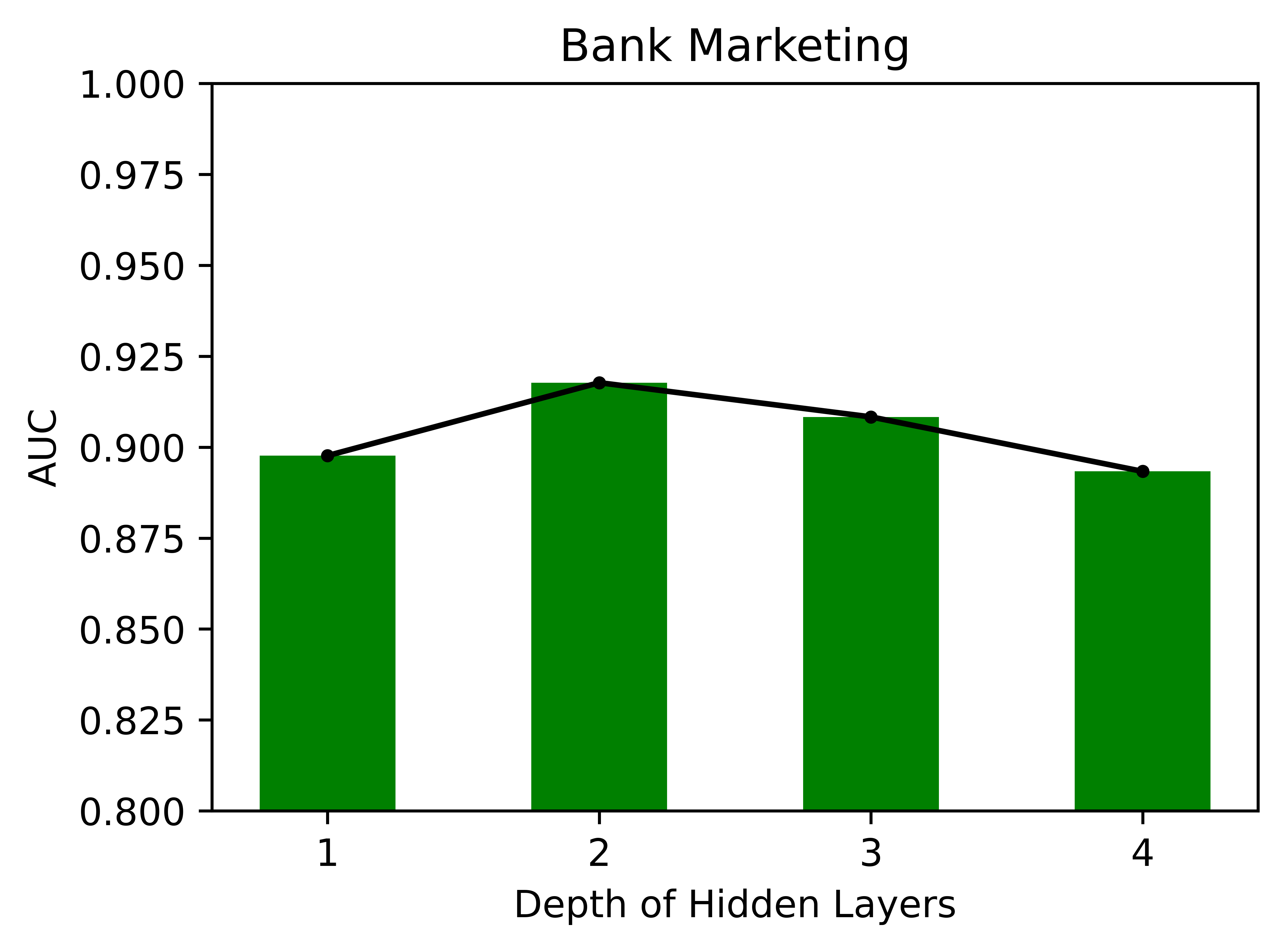}
\end{minipage}
\begin{minipage}[t]{0.32\textwidth}
\centering
\includegraphics[width=\textwidth]{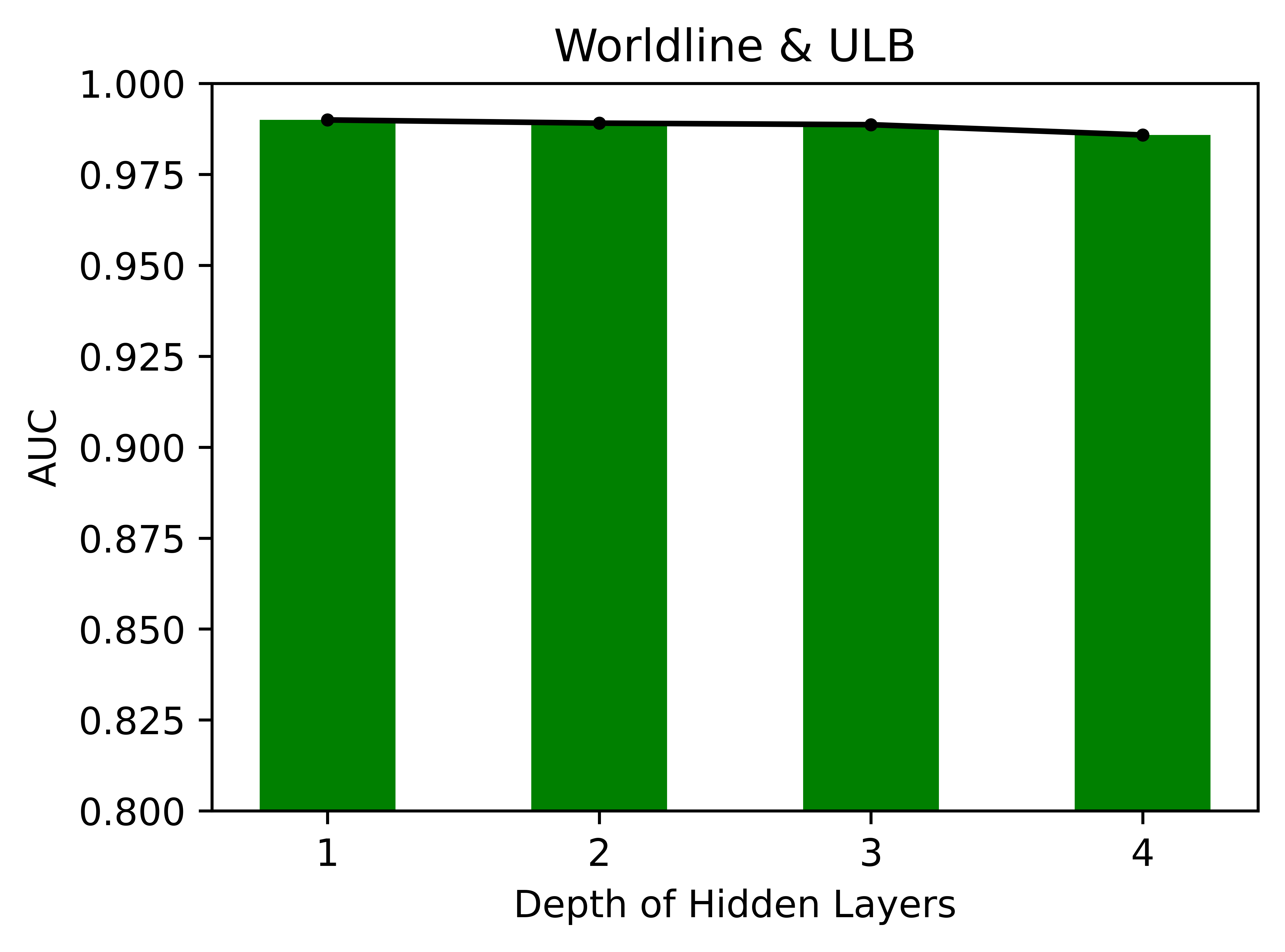}
\end{minipage}
\\
\begin{minipage}[t]{0.32\textwidth}
\centering
\includegraphics[width=\textwidth]{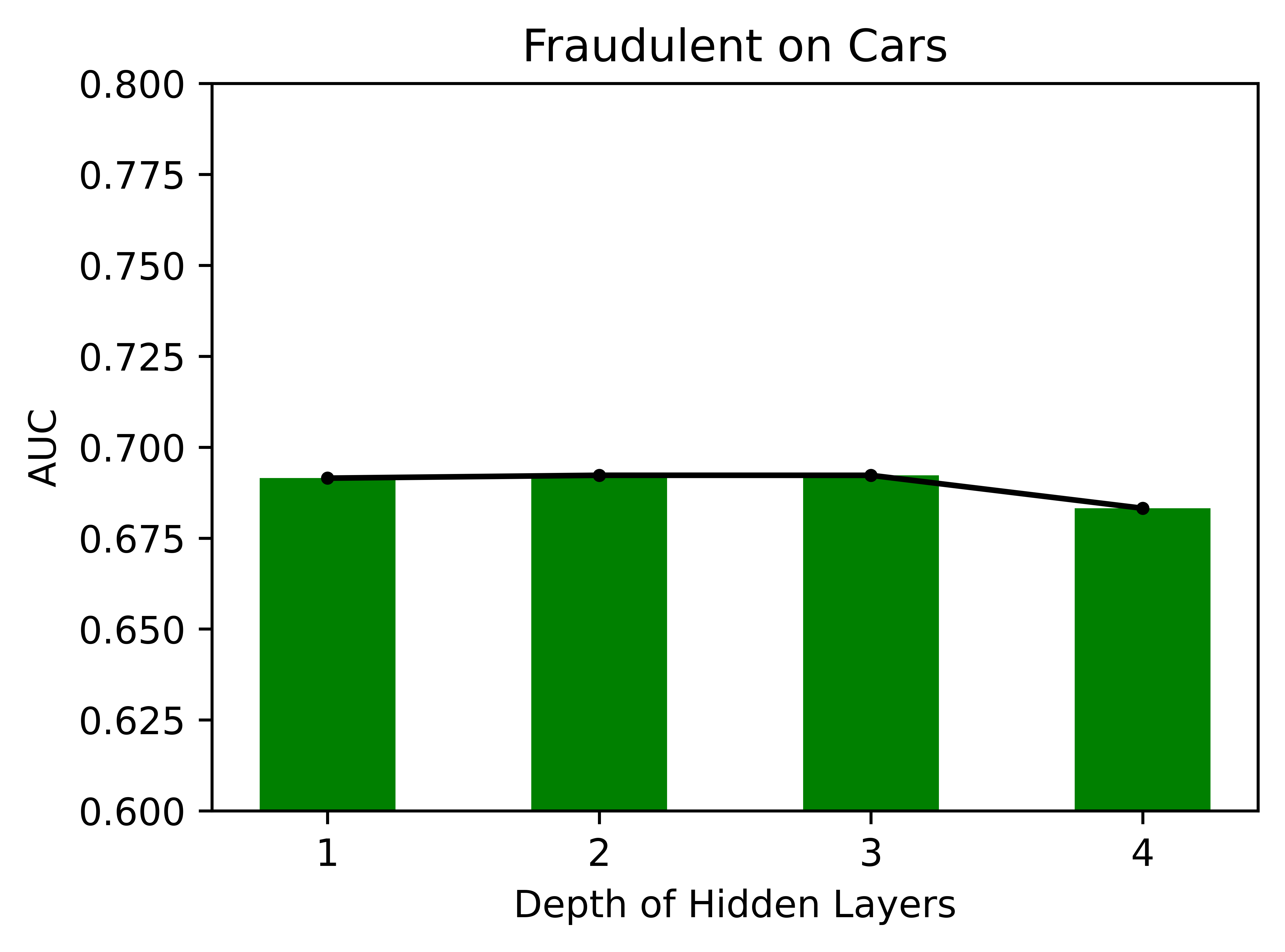}
\end{minipage}
\begin{minipage}[t]{0.32\textwidth}
\centering
\includegraphics[width=\textwidth]{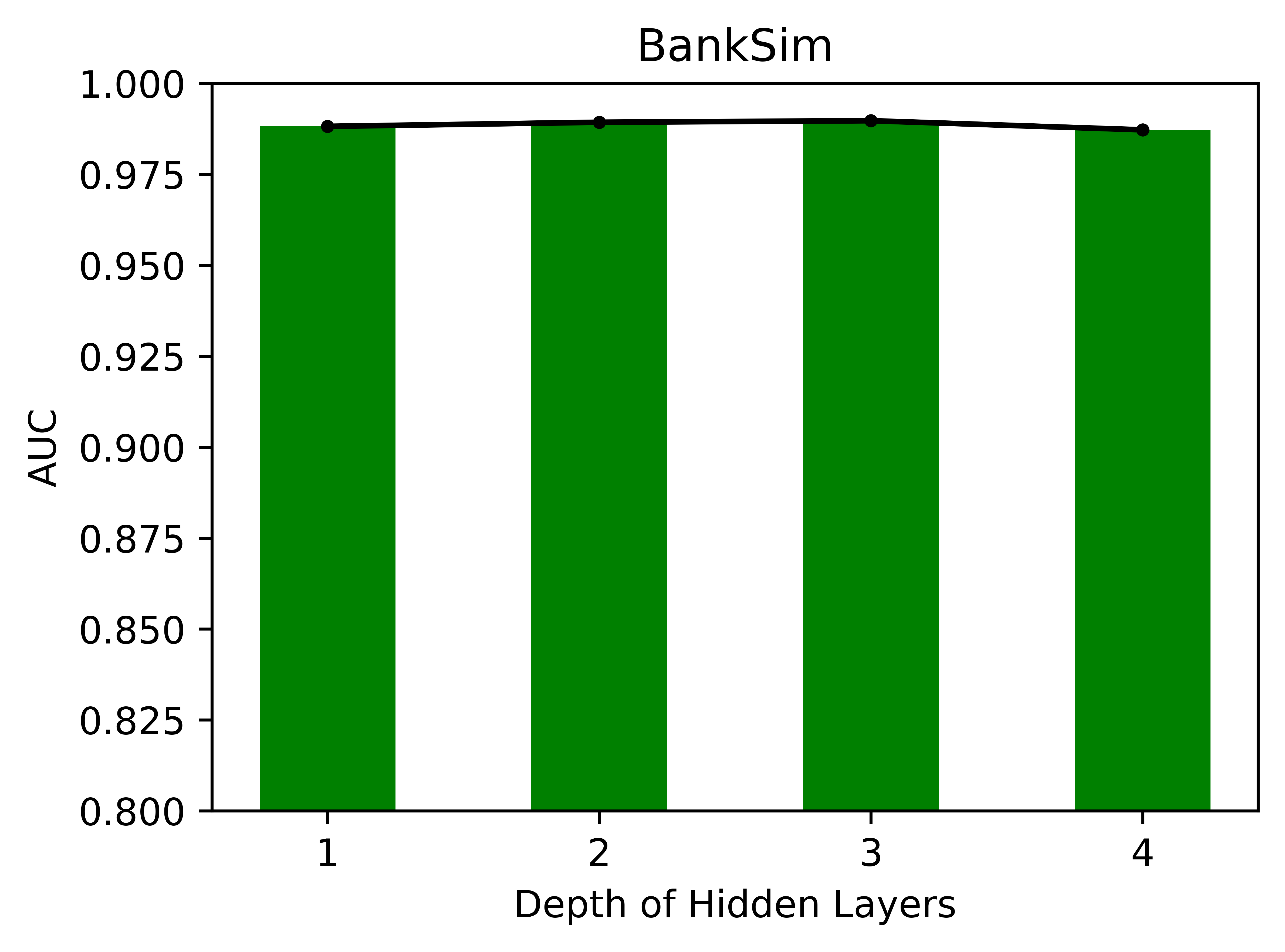}
\end{minipage}
\caption{The effect of the depth of the neural networks in DBDT.}\label{layer}
\end{figure}

Figure \ref{layer} shows the influence of the depth of the neural networks in DBDT. With the deeper hidden layers, the model performance metrics has a significant improvement, then basically reaches the maximum and there is a downward trend finally. This illustrates complex networks could significantly improve the model's fraud detection capabilities. The downward trend could be caused by the networks structure. Therefore, the feature representation for fraud detection is east relatively because of the limited feature dimensions and a shallow multilayer network can already achieve good results. If it is necessary or feature learning is difficult, we can use a 3-layer neural network, but this will multiply the model complexity.
\vspace{-2mm}
\subsection{Case Study}

One of the criticism of deep learning methods is that the interpretability is poor, so we
often call neural network the black box model. In many specific applications, the expectations
of business operators for deep learning include not only the higher prediction accuracy, but
also the better interpretability. Taking fraud detection as an example, operators not only
care about the AUC of the model, but more hope that the model can answer questions such
as why is this person more likely to cheat than others. Therefore, we hope to take into
account both the generalization performance and interpretability of DBDT, and try to open
the model in terms of feature engineering.

We make predictions on Load Data, a publicly available data from LendingClub.com. This dataset represents 9,578 3-year loans that were funded through the LendingClub.com platform from May 2007 to February 2010. We have 1533 frauds out of those 9578 transactions (here fraud means that the loan was not paid back in full), and the positive class (frauds) accounts for
16.01\% of all transactions. Since we embed the neural network into the decision tree node, the feature engineering cannot be calculated using the conventional method such as Gini coefficient and information gain. Hence, we use OOB (Out of Bag) method to calculate the feature importance. First, we use the trained model to predict the OOB data, and calculate the scores of performance metrics such as AUC; then, we randomly scatter each feature in the OOB data one by one, recalculate the performance metrics, and evaluate the feature importance according to the rate of change of the score.

\begin{figure}[t]
\centering
\includegraphics[height=2in]{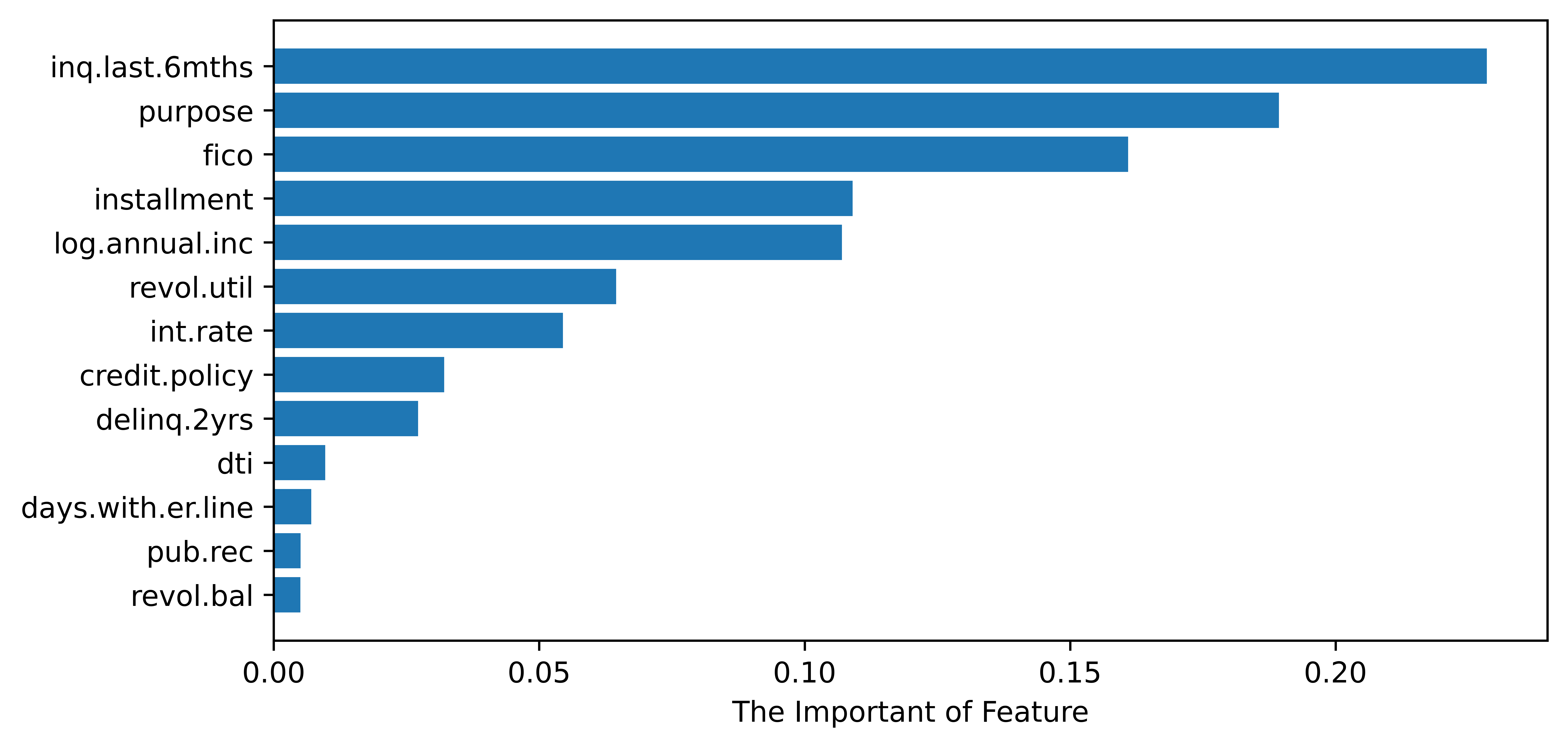}
\caption{Feature importance ranking.} \label{feature8}
\end{figure}

We use the AUC as scoring criterion and rank feature importance as above. The results are shown in Figure \ref{feature8}, from which we find that the top 5 important features are (1) The number of times the borrower's credit has been inquired in the past 6 months (inq.last.6mths, 22.8\%), (2) Loan purpose (purpose, 18.9\%), (3) Borrower's FICO score (fico, 16.1\%), (4) Borrow's installment (installment, 10.9\%) and (5) The natural logarithm of borrower's annual income (log.annual.inc, 10.7\%). According to the ranking of feature importance, the most important feature is not the personal credit and assets, but the number of times the borrower's credit has been inquired (inq.last.6mths), which reflects the borrower's frequency of loan activity over six-month period. One possible explanation is that the more frequent to borrow, the more likely to default. The second-ranked feature is the purpose of borrowing, which is reasonable for that the repayment situation of different purposes of loans has obvious differences. DBDT has learned the difference, and thus the importance of this feature is very high. The next three features reflect the borrower's personal credit and personal property. Borrowers with good credit and income are more likely to repay, which is in line with people's perceptions. 

\end{spacing}
\vspace{-4mm}
\section{Conclusion}
\begin{spacing}{2.0}
\vspace{-2mm}
In this paper, we propose a novel fraud detection approach named deep boosting decision trees (DBDT). DBDT embeds neural networks into gradient boosting  to improve its representation learning capability and meanwhile maintain the interpretability, and furthermore, DBDT can be trained by a compositional AUC maximization approach to deal with data imbalances at algorithm level. Extensive experiments on a large set of real-life fraud detection datasets 
confirm the highly competitive nature of DBDT in terms of two dimensions it aims to reconcile: predictive accuracy and interpretability. 

Specifically, results demonstrate that when trained using SGD without considering data imbalances, DBDT-SGD significantly outperforms the general classification algorithms such as AdaBoost and GBDT but not as good as re-sampling based approaches, from  which we can draw two conclusions: (1) DBDT-SGD can act as a general machine learning algorithm with superior performance to the popular methods; (2) some means of dealing with data imbalances is necessary in fraud detection task. Furthermore, it was found that when trained with the compositional AUC maximization approach, DBDT achieved superior performance to those re-sampling based approaches, which illustrates that: (1) DBDT with compositional AUC maximization strategy provides an effective means for fraud detection; (2) the proposed compositional AUC maximization strategy can effectively improve the adaptability of the algorithm to data imbalance, and thus has the potential to be extended to other algorithms. Moreover, experiments on  parameter sensitivity show that DBDT is relatively stable for the parameters, and model interpretability of DBDT was assessed in detail by means of a case study, investigating the fraud prediction  in the setting of a loans data, which confirms that DBDT maintains good interpretability.

In summary, the contributions of this study are thus the following: (1) we combine neural networks and gradient boosting to get an end-to-end structure (DBDT) with good interpretability, and demonstrate its superiority to the general classification algorithms; (2) we employ a compositional AUC maximization approach to train the DBDT model, which can effectively deal with the data imbalances in fraud detection and outperform the state-of-the-arts; (3) an extensive benchmark study is conducted to compare the performance of DBDT to the well-established competing algorithms; (4) a case study is conducted to illustrate the good interpretability of DBDT as an end-to-end model.

The value of DBDT is intriguing and its usefulness could be further explored in future research. On one hand, it is not hard to see that DBDT can be treated as a general framework to embed neural networks into tree based machine learning approaches, thus we can try some other popular boosting algorithms such as XGBoost, LightGBM to get better performance. On the other hand, the proposed compositional AUC maximization strategy is proven to be an effective way to deal with the data imbalance problem at algorithm level, and can be extended to other algorithms to improve the adaptability to imbalanced data.
\end{spacing}

\vspace{-6mm}
 \begin{spacing}{1.5}
 \bibliographystyle{elsarticle-num} 
 \bibliography{dss}
 \end{spacing}





\end{document}